\pdfoutput=1
\documentclass[11pt]{article}
\usepackage{newunicodechar}
\newunicodechar{∞}{\ensuremath{\infty}}
\usepackage[preprint]{acl}

\usepackage{times}
\usepackage{latexsym}
\usepackage{graphicx}
\usepackage[T1]{fontenc}
\usepackage[utf8]{inputenc}

\usepackage{microtype}

\usepackage{inconsolata}

\usepackage{graphicx}

\usepackage[utf8]{inputenc} % allow utf-8 input
\usepackage[T1]{fontenc}    % use 8-bit T1 fonts
\usepackage{url}            % simple URL typesetting
\usepackage{booktabs}       % professional-quality tables
\usepackage{amsfonts}       % blackboard math symbols
\usepackage{nicefrac}       % compact symbols for 1/2, etc.
\usepackage{microtype}      % microtypography
\usepackage{xcolor}         % colors
\usepackage{multirow}
\usepackage{graphicx}
\usepackage{pifont}

\newcommand{\fref}[1]{Figure~\ref{#1}}
\newcommand{\tref}[1]{Table~\ref{#1}}
\newcommand{\sref}[1]{\S\ref{#1}}

\usepackage{microtype}
\usepackage{graphicx}
\usepackage{amsmath}
\usepackage{bbm}
\usepackage{tcolorbox}

\usepackage{enumitem}
\usepackage{CJKutf8}

\usepackage[]{todonotes}

\title{Ref-Long: Benchmarking the Long-context Referencing Capability of Long-context Language Models}

\newcommand{\authorsep}{\quad}

\author{
Junjie Wu$^1$\thanks{Equal contribution.}\authorsep
Gefei Gu$^2$\footnotemark[1]\authorsep
Yanan Zheng$^3$\authorsep
Dit-Yan Yeung$^1$\authorsep
Arman Cohan$^3$\authorsep
\\
\textsuperscript{1}Hong Kong University of Science and Technology\\
\textsuperscript{2}Carnegie Mellon University\\
\textsuperscript{3}Yale University\\
\texttt{junjie.wu@connect.ust.hk} \quad \texttt{gefeig@andrew.cmu.edu} \\ \texttt{\{yanan.zheng, arman.cohan\}@yale.edu} \quad \texttt{dyyeung@ust.hk}
}
\begin{document}
\maketitle
\begin{abstract}
Long-context language models (LCLMs) have exhibited impressive capabilities in long-context understanding tasks. Among these, long-context referencing—a crucial task that requires LCLMs to attribute items of interest to specific parts of long-context data—remains underexplored.
%However, existing long-context benchmarks face several limitations: they are often biased, lack cost-effectiveness, or fail to provide sufficient challenges for LCLMs.
%\ac{is this true? I feel this is a strong statement. instead we can frmae it as: we focus on an underexplored aspect of long-context understanding which is referencing.}
To bridge this gap, this paper proposes \textbf{Ref}erencing Evaluation for \textbf{Long}-context Language Models (\textbf{Ref-Long}), a novel benchmark designed to assess the long-context referencing capability of LCLMs. Specifically, Ref-Long requires LCLMs to identify the indexes of documents that reference a specific key, emphasizing contextual relationships between the key and the documents over simple retrieval.
%\ac{how is this different than retrieval?} 
Based on the task design, we construct three subsets ranging from synthetic to realistic scenarios to form the Ref-Long benchmark. Experimental results of 13 LCLMs reveal significant shortcomings in long-context referencing, even among advanced models like GPT-4o. To further investigate these challenges, we conduct comprehensive analyses, including human evaluations, task format adjustments, fine-tuning experiments, and error analyses, leading to several key insights. Our data and code can
be found in \url{https://github.com/wujunjie1998/Ref-Long}.
%\ac{problem formulation isn't very clear. motivation is weak. try to rewrite to address these.}
\end{abstract}

\section{Introduction}
\label{sec:introduction}

\begin{figure}
    \centering
    \includegraphics[width=0.48\textwidth]{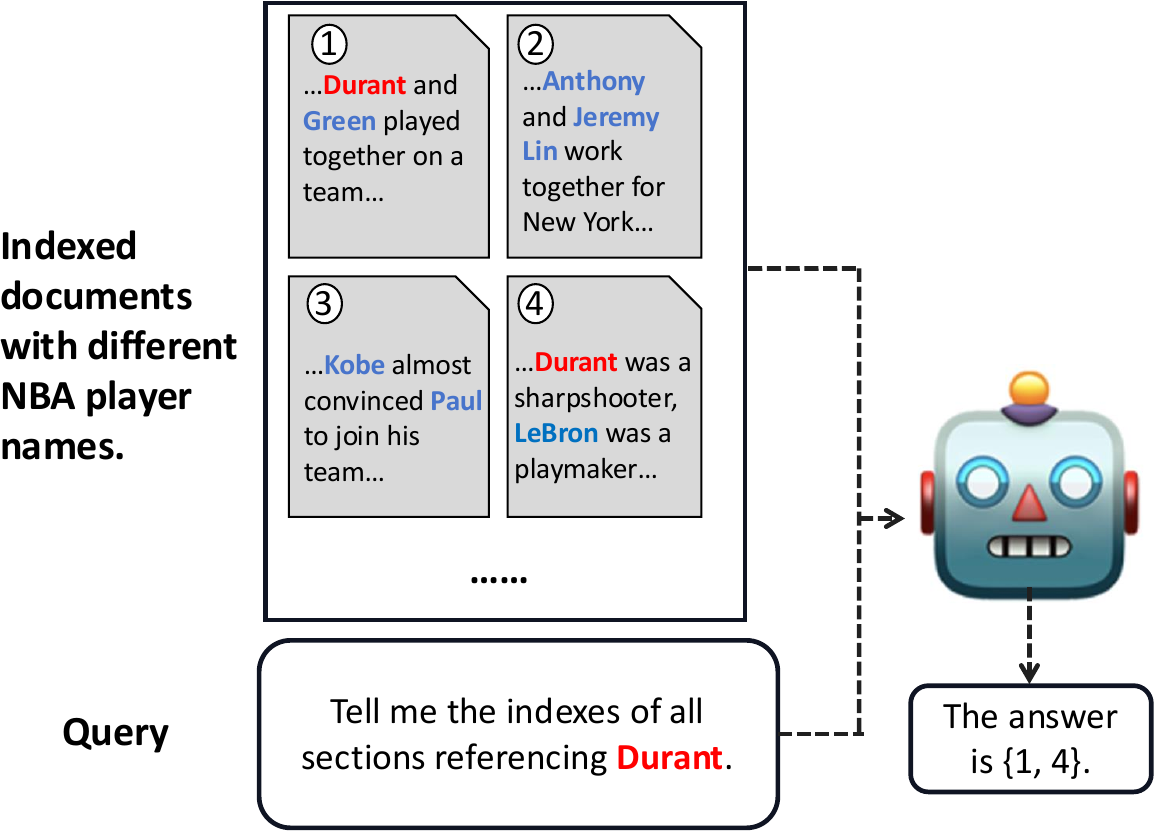}
    \caption{An example Ref-Long task. Given a long-context input with indexed documents that include several NBA players names, an LCLM is asked to give the indexes of documents that reference ``Durant'' (marked as red). Names other than ``Durant'' are marked as blue. %\yn{Do we need to change the query to a more practical one? I mean, the key implicitly appears in the query.}
    } 
    %\vspace{-0.2in}
    \label{fig:overview}
\end{figure}

%The long-context understanding capability is crucial for 
Long-context language models (LCLMs) have demonstrated remarkable long-context capabilities in tasks such as multi-document question answering~\cite{bai2024longbench, wang-etal-2024-leave} and %multi-document 
summarization~\cite{QuerySum, laban2024summary}. Among long-context capabilities, \textbf{long-context referencing}, referring to LCLMs' ability to correctly attribute interested items to specific parts of extensive long documents,
%Some work uses the term \textit{citation} specifically to mean citing papers. To avoid confusion, we use the term \textit{referencing} for a broader concept: providing document-level information related to a given interested item, i.e., query. Under this definition, citing papers becomes a specialized form of long-context referencing.
%\ac{I think (if you are talking about the same thing) ``attribution'' is more commonly used in the literature than referencing. But what we are doing is maybe `fine-grained attribution' where we want to point to specific parts within a long doc, rather than pointing to a specific doc. }
is crucial and has many real-world applications.~\footnote{The term \textit{referencing} differs from \textit{retrieval} in that it requires LCLMs to not only retrieve keys from long context, but also know the location (specific parts) where these keys appears in the long context.}
For instance, legal practitioners need to quickly identify the specific chapter within the relevant legal code when presented with a particular case or provision, and financial professionals need to swiftly determine which financial report contains the given data.

%evidence from extensive long documents when generating an answer, 
%By providing appropriate references, users can trace the source documents to validate the faithfulness of generated answers, significantly enhancing the reliability and trustworthiness of long-context large language models (LCLMs)~\cite{TrustLLM}. \jw{maybe add some real-world applications of long-context referencing here.}

Although various benchmarks exist for evaluating the long-context capabilities of LCLMs, very few assess the dimension of long-context referencing. Existing long-context benchmarks can be broadly categorized into two types. On one hand, \textit{general long-context benchmarks}, such as LongBench~\cite{bai2024longbench}, L-Eval~\cite{an-etal-2024-l}, %Marathon~\cite{DBLP:conf/acl/ZhangLLYLCLY24}
NOCHA~\cite{karpinska2024one}, and a combination of them (HELMET)~\cite{yen2024helmet}, are either synthesized by adding irrelevant texts into short-context NLP tasks, which results in unrealistic context distributions and biased evaluations, or constructed from scratch with human annotations, which requires substantial resources and complicated human efforts. %\yn{please check previous sentence change!}
% are either synthesized by adding irrelevant text to short-context NLP tasks, resulting in unrealistic context distributions and biased evaluations, or require substantial resources to construct from scratch. 
% Also, the long input lengths of these benchmarks make them challenging for humans as well, further complicating efforts to estimate their difficulty levels.
On the other hand, as a specific and well-studied type of long-context benchmark, \textit{retrieval-based benchmarks} such as Needle-in-a-Haystack \cite{needle}, %NeedleBench \cite{li2024needlebench}, 
Counting-stars \cite{song2024counting}, and RULER~\cite{hsieh2024ruler}, focus on matching and retrieving target texts but often overlook the nuanced relationships between the retrieved texts and their surrounding contexts. This makes these benchmarks overly simplistic and not comprehensive. Moreover, while existing benchmarks address some aspects of long-context understanding, they fail to effectively evaluate long-context referencing, highlighting the urgent need for practical and robust benchmarks in this area.
%\ac{How about realistic long-context benchmarks such as HELMET?}

To address the above issues, this work proposes a novel benchmark called \textbf{Ref}erencing Evaluation for \textbf{Long}-context Language Models (\textbf{Ref-Long}), which is specifically designed to assess the long-context referencing capability of LCLMs. As illustrated in~\fref{fig:overview}, given several indexed long documents and a query that includes ``Durant'', LCLMs are required to  %not only answers but also the corresponding supporting documents simultaneously. \textcolor{red}{The benchmark highlights in its setting design---xxxxx}.
not only identify ``Durant'' in the given documents, but also need to figure out the indexes of documents that reference ``Durant'' rather than other NBA players. 
This task setting has several advantages. %1) it is built on coherent contexts and reflects realistic distributions; it is cost-efficient and manageable for human annotators; and it considers the relationships between key information and distracting information within long contexts. 
First, it considers the relationship information between the specific key and its surrounding context, which forces LCLMs to genuinely understand long contexts instead of simply relying on shortcuts to retrieve the key. As a result,~\sref{sec:fluent with abrupt} and~\sref{sec:realistic} show that Ref-Long presents certain level of difficulty that challenges even the most advanced LCLMs (e.g., GPT-4o \cite{gpt4o}). Second, Ref-Long tasks can be constructed cost-efficiently, as only the locations of specific keys are required. Furthermore, as shown in~\sref{sec:human}, Ref-Long tasks remain manageable for human annotators, allowing their difficulty level to be estimated. %This comparison between humans and LCLMs' performances highlights the importance of addressing LCLMs' limitations in long-context referencing.
%To better understand the challenges posed by our Ref-Long, we conduct a series of in-depth investigations. They examine \textcolor{red}{xxxxxx}

Following the task setting, we construct three subsets, ranging from synthetic to realistic scenarios, to form the Ref-Long benchmark and evaluate 13 LCLMs. Experimental results on these subsets reveal that all LCLMs struggle with Ref-Long tasks (\sref{sec:fluent with abrupt}, \sref{sec:realistic}), highlighting their lack of long-context referencing capability. Furthermore, we investigate the challenge faced by LCLMs from several perspectives. Motivated by findings in~\cite{wu-etal-2025-understanding,yu-etal-2025-stochastic} that LLMs may struggle with tasks easily handled by humans, we first conduct a human evaluation in~\sref{sec:human} to assess whether task difficulty contributes to the challenges in Ref-Long.
 Next, we examine if the issue comes from the format of task queries by applying alternative formats to evaluate LCLMs (\sref{sec:factors}). Additionally, we explore whether fine-tuning can mitigate LCLMs' limitations in long-context referencing (\sref{sec:fine-tuning}). Finally, we perform error analysis on LCLMs' failed cases (\sref{sec:error}).  In summary, our contributions are threefold.
\begin{enumerate}[noitemsep,nolistsep,leftmargin=*]
    \item First, we introduce Ref-Long, a novel benchmark that solves long-context referencing and limitations in existing long-context benchmarks.
    \item Second, we demonstrate Ref-Long is uniquely challenging for state-of-the-art LCLMs yet remains accessible for human annotators, underscoring its importance for advancing the field.
    \item Finally, through comprehensive analyses, we identify several findings that could be used to facilitate LCLMs' long-context referencing and understanding capabilities.
\end{enumerate}
%Our work marks a significant step forward in understanding and improving long-context capability, paving the way for future research to develop more reliable and faithful long-context LLMs. 
\section{Related Works}

With the rapid evolution of LLMs, several benchmarks have been proposed to evaluate their long-context understanding capabilities, which can be categorized into two types. 

\paragraph{General Long-Context Benchmarks.}
The first type extends general short-context tasks to long-context scenarios, which provides comprehensive evaluations of LCLMs. %~\cite{bai2023longbench, zhang2023marathon, an-etal-2024-l, zhang2024bench, levy2024same, karpinska2024one, ma2024mmlongbench, laban2024summary}. 
However, several issues exist in these works. On one hand,~\cite{bai2024longbench, levy2024same} build their benchmarks by directly appending external context around a short NLP task, which is not realistic and may introduce additional bias to LCLMs as the token distribution between the task and the newly added context may differ significantly. On the other hand,~\cite{bao2021conversations,yu2023personality, zhang2024marathon,an-etal-2024-l, zhang2024bench, xu2024fine, karpinska2024one, mammlongbench, laban2024summary, yen2024helmet} build NLP tasks like question answering (QA) and summarization on long-context data from scratch. However, this approach is costly and requires significant human effort. For instance,~\citet{karpinska2024one} spent 3330 USD to annotate just 1001 QA pairs on novels.

\paragraph{Retrieval-based Long-Context Benchmarks.} Another category of benchmarks, derived from the needle-in-a-haystack task~\cite{kamradt2023needle}, evaluates LCLMs' ability to retrieve specific keys or supporting facts from long-context data~\cite{li2024needlebench, song2024counting, zhang2024bench, kuratov2024babilong, wang-etal-2024-leave, vodrahalli2024michelangelo, robertsneedle, hsieh2024ruler}. However, these approaches have several limitations. First, they only require LCLMs to retrieve keys from documents without considering the relationships between keys and their surrounding context, making the tasks relatively simple. For instance, GPT-4o achieves over 70\% accuracy on most benchmarks even with input lengths of 128K tokens. Additionally, if an LCLM retrieves keys successfully for subsequent tasks, these tasks effectively become short-context tasks.
%\yn{Currently, both the general and retrieval benchmark parts have no differentiated information as it appeared in the Introduction. Shall we make them more concrete? Or try to summarize those sentences in Intro to make them succinct. }

%\yn{Shall we add another part: Citation based Benchmark. We consider this type of works as a type of similar works, but we  distinguish ourself with them.}

To address these issues, we propose Ref-Long, an effective benchmark that evaluates the referencing ability of LCLMs. While~\cite{laban2024summary, wang-etal-2024-leave} also include referencing tasks, their studies lack systematic analysis and are constrained by limited task settings. Moreover,~\cite{gao2023enabling,bai2024longcite,tang2024citeeval} evaluate LCLMs by requiring them to generate sentence-level references alongside answers in QA tasks. This setup differs from Ref-Long, which requires broader contextual referencing around a key idea rather than isolated chunks linked to a specific question. Also, this setup is less generalizable, as it depends on either sentence-level ground-truths (requiring extra annotation) or LLMs as external evaluators for further evaluation.

\begin{table*}[tb]
\small
\centering
\setlength{\tabcolsep}{4.6mm}
\begin{tabular}{ll|cc|cc|cc}
\toprule
\multirow{2}{*}{\textbf{LCLM}} & & \multicolumn{2}{|c}{\textbf{Easy}} & \multicolumn{2}{c}{\textbf{Medium}} & \multicolumn{2}{c}{\textbf{Hard}} \\
\cmidrule(lr){3-4} \cmidrule(lr){5-6} \cmidrule(lr){7-8}
& & \textbf{F1}$\uparrow$ & \textbf{Ex Acc}$\uparrow$ & \textbf{F1}$\uparrow$ & \textbf{Ex Acc}$\uparrow$ & \textbf{F1}$\uparrow$ & \textbf{Ex Acc}$\uparrow$ \\
\midrule
ProLong-8B-64K & & 16.44 & 0.00 & 19.59 & 0.00 & 38.11 & 0.00 \\
ProLong-8B-512K &  & 20.02 & 0.00 & 24.82 & 0.00 & 40.06 & 0.00 \\
LongCite-8B & & 6.42 & 1.00 & 5.74 & 1.00 & 10.20 & 0.00 \\
Llama-3.1-Ins-8B &  & 30.95 & 2.00 & 30.00 & 0.00 & 38.85 & 0.00 \\
Phi-3-mini & & 23.04 & 8.00 & 20.82 & 5.00 & 21.82 & 0.00 \\
Qwen2.5-Ins-7B & & 30.46 & 13.00 & 24.63 & 8.00 & 20.37 & 0.00 \\
%\midrule[0.5pt]
Qwen2.5-Ins-72B & & 73.09 & 39.00 & 70.77 & 22.00 & 60.90 & 5.00 \\
Llama-3.1-Ins-70B & & 74.47 & 41.00 & 64.43 & 19.00 & 52.21 & 4.00 \\
Llama-3.3-Ins-70B & & 76.52 & 43.00 & 66.92 & 19.00 & 56.23 & 4.00 \\
Gemini-1.5-Flash & & 82.16 & 51.00 & 74.19 & 29.00 & 64.65 & 2.00 \\
GPT-4o mini & & 87.69 & 67.00 & 85.49 & 52.00 & 68.64 & 7.00 \\
%\midrule[0.5pt]
Gemini-1.5-Pro & & 89.29 & 67.00 & 80.20 & 44.00 & 65.24 & 9.00 \\
GPT-4o & & \textbf{93.45} & \textbf{75.00} & \textbf{90.35} & \textbf{61.00} & \textbf{75.38} & \textbf{19.00} \\
\bottomrule
\end{tabular}
\caption{Results on Ref-Long-A, where Ex Acc is in percentage and ``Ins'' means ``Instruct''. 
The best results under each column are \textbf{boldfaced}. For conciseness, we only list the 24K results under Multi here, and show Single results and 8K/16K results under Multi in~\tref{tab:complete setting1}. %Due to the budget limit, we only run Gemini-1.5-Pro on Multi since Single is not challenging for models in its size.
}
%\vspace{-0.2in}
\label{tab:setting1}
\end{table*}

\section{The Ref-Long Benchmark}
\label{sec:benchmark}

To overcome the issues of existing long-context benchmarks, we introduce Ref-Long, a new benchmark that aims to assess the long-context referencing capability of LCLMs. %which extends beyond the original needle-in-a-haystack task. 
Specifically, Ref-Long challenges LCLMs to not only retrieve a specific key from a collection of documents but also identify the indexes of all documents that reference the key. Following this task setup, Ref-Long includes three distinct sub-datasets that encompass both synthetic and real-world data (\sref{sec:fluent with abrupt}, \sref{sec:fluent with fluent}, \sref{sec:paper citation}), enabling a systematic evaluation of the referencing capabilities of LCLMs.

\subsection{Task Setup}
\label{sec:task setup}

Generally, Ref-Long tasks are built upon a candidate set of documents, where each document contains $\boldsymbol{N}$ distinct keys and keys can be overlapped across documents. To create an Ref-Long task, we first randomly select $\boldsymbol{M}$ documents from the candidate set and index them with numbers, then sample a specific key $\boldsymbol{k}$ that appears in these $\boldsymbol{M}$ documents. The task of an evaluated LCLM is to identify the indexes of all documents within the $\boldsymbol{M}$ documents that referencing $\boldsymbol{k}$. The overall process is illustrated in~\fref{fig:overview}.

\subsection{Evaluation Metrics}
\label{sec:metrics}

We evaluate LCLMs' performance on Ref-Long tasks using exact match accuracy (\textbf{Ex Acc}), which measures whether the output indexes match the ground truth exactly (ignoring order). A score of 1 is assigned only if the indexes are an exact match. Additionally, we compute the F1 score (\textbf{F1}) of the precision and recall, giving equal weight to both, as a supplement. For both metrics, higher scores refer to better performance. 

\subsection{Evaluated LCLMs}
\label{sec:evaluated lclms}

We evaluate both closed-source and open-source LCLMs. For closed-source LCLMs, we include the powerful GPT-4o~\cite{openai2024gpt4o} and Gemini-1.5-Pro~\cite{geminiteam2024gemini} along with their smaller variants, GPT-4o Mini and Gemini-1.5-Flash. As for open-source LCLMs, we evaluate the following: Llama-3.1-Instruct (70B, 8B)~\cite{llama3.1modelcard}, Llama-3.3-Instruct-70B, Qwen2.5-Instruct (72B, 7B)~\cite{qwen2.5}, Phi-3-mini (\texttt{Phi-3-mini-128k-instruct})~\cite{abdin2024phi3}. We also include Prolong-8B-64K/512K (\texttt{Llama-3-8B-ProLong-Instruct}) that have been elaborately fine-tuned on long-context data~\cite{gao2024train}, and LongCite-8B (\texttt{LongCite-llama3.1-8B})~\cite{zhang2024longcite} that is fine-tuned on QA responses and sentence-level references for comparison. %All the evaluated LCLMs, except for ProLong-8B-64K, support context sizes exceeding 128K tokens and are capable for the following experiments. Due to computational resource constraints, we use the 4-bit versions of Llama-3.1-Instruct-70B and Qwen2.5-72B. 
Further details regarding the inference setting of these LCLMs are provided in Appendix~\sref{appendix:inference details}.

\section{Abrupt Key in Fluent Context}
\label{sec:fluent with abrupt}

\subsection{Dataset}
\label{sec:setting1 data}
We start from crafting a synthetic dataset named \textbf{Ref-Long-A}brupt (\textbf{Ref-Long-A}) from scratch following the task setup in~\sref{sec:task setup} as the first subset of Ref-Long. %aiming to provide an initial insight into the performance of LCLMs on long-context referencing tasks. 
Specifically, we extract all essay files from the Paul Graham Essays English context dataset used in~\citet{needle}, and randomly concatenate and truncate these files %into a single text. We then truncate from the start of this concatenated text 
to generate 100 documents with approximately 1,000 tokens each. Next, %we construct and insert keys into the 100 generated documents. 
we randomly select 1 (\textbf{Single}) /5 (\textbf{Multi}) positions in each document to insert a template sentence: "\textit{The little penguin counted \{$\boldsymbol{num}$\} \ding{72}}" that is abrupt and irrelevant to the background context\footnote{This template follows~\citet{song2024counting}, while they only evaluates key retrieval and thus not challenging for LCLMs.}, where the integer $\boldsymbol{num}$ $\in [a, b)$ refers to the number of stars the little penguin counted and forms the keys that we are interested in. When crafting an Ref-Long task, we first randomly sample $\boldsymbol{M}$ documents, then select a $\boldsymbol{num}$ that appears in these documents as the specific key for evaluation.
%and the $\boldsymbol{m}$ numbers are unique within each document, and the key $\boldsymbol{k}$ is also picked from these numbers.
%Intuitively, the difficulty of an Ref-Long task depends on two factors: the input length of the task and the frequency of the specific key $\boldsymbol{k}$ in the sampled $\boldsymbol{M}$ documents. While input length is mainly controlled by $\boldsymbol{M}$, aspects like number of keys in each document and the range of the candidate keys can both affect the second factor. 

%Since the above crafting process is automatic and efficient, it enables us to adjust different parameters 
For a comprehensive evaluation, %Specifically, we set $\boldsymbol{m}$ to 1 (\textbf{Single}) and 5 (\textbf{Multi}) where larger $\boldsymbol{m}$ leads to higher frequency of all the keys. 
%under each value of $\boldsymbol{m}$, 
we set the range of integer $\boldsymbol{num}$ to $[0, 100)$ (\textbf{Easy}), $[0, 60)$ (\textbf{Medium}) and $[0, 20)$ (\textbf{Hard}) under Single and Multi to form 6 settings, where a smaller range leads to higher frequency of different $\boldsymbol{num}$ across documents. %For each combination of $\boldsymbol{m}$ and $[a, b)$, 
$\boldsymbol{M}$ is set to \{8, 16, 24\} under each setting, correspond to input lengths of 8K, 16K, and 24K tokens. For each input length, we generate 100 Ref-Long tasks in an aggregative way (e.g., the first 8 documents and the specific key in a 16K task is the same as the corresponding 8K task) to reduce the effect of randomness, and finally forms the Ref-Long-A subset with 1800 distinct tasks. We evaluate all the LCLMs in~\sref{sec:evaluated lclms} on Ref-Long-A and list the 24K results under Multi in~\tref{tab:setting1} (check Appendix~\sref{appendix:prompts} for the prompts we use).

\subsection{Results}
\label{sec:setting1 results}

We observe that %on the easiest setting (Single-Easy), smaller LCLMs achieve results comparable to larger models. 
larger models consistently perform better on the Ref-Long-A tasks, aligning with expectations that models with more parameters could have stronger capabilities, which validates the reliability of Ref-Long's conclusions. %Among open-source LCLMs, we find that Llama-Ins-70B outperforms Qwen2.5-Ins-72B, while Qwen2.5-Ins-7B leads among small-size LCLMs. 
When comparing the two 70B Llama models, Llama-3.3-Ins-70B surpasses Llama-3.1-Ins-70B, suggesting that pre-training on more multilingual data enhances referencing capabilities. However, since the improvements under the Multi setting are limited, we only include the Llama-3.1-Ins-70B version in the rest of the experiments. 

Surprisingly, even the strongest LCLMs faces significant challenges on the multi-hard setting, where the context length is just 24K—far below their maximum context size(e.g., the highest Acc score under the multi-hard setting is only 19.00). \textbf{This surprising result demonstrates that current LCLMs lack the capability to grasp the positional relationships between keys and the contexts in which those keys are retrieved, a capability that is essential for effective long-context understanding}. Also, we notice that models elaborately fine-tuned on long-context data do not obtain competitive results on Ref-Long tasks, and further investigate this observation in~\sref{sec:fine-tuning}.

\paragraph{Ablation Study on Input Length.} 
\begin{figure*}
    \centering
    \includegraphics[width=0.94\textwidth]{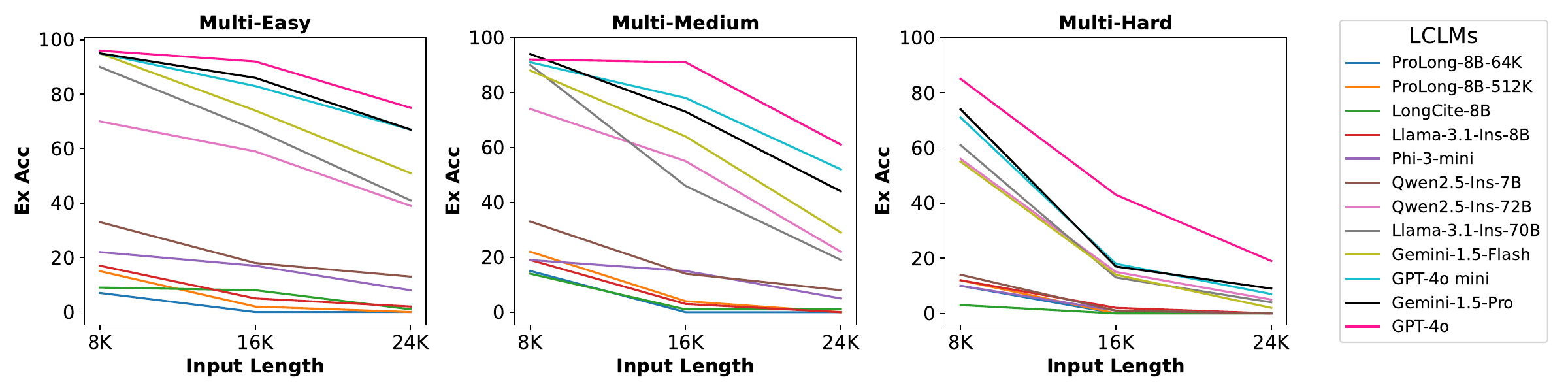}
    \caption{LCLMs' performances drop consistently as input length grows. See~\tref{tab:complete setting1} for numerical results.
} 
%\vspace{-0.2in}
    \label{fig:ablation}
\end{figure*}
As a benchmark for long-context evaluation, it is essential to ensure that the difficulty of Ref-Long tasks increases with input length; otherwise, extending the input length would be meaningless. To verify this, we plot the Exact Acc scores for Multi-Easy, Multi-Medium, and Multi-Hard tasks with input lengths of 8K, 16K, and 24K tokens across all LCLMs. The results are shown in~\fref{fig:ablation}. As shown, the performance of all LCLMs consistently drops as the input length increases, further demonstrating the reliability of Ref-Long for long-context evaluation.

%\paragraph{Error Analysis.} It's observed from~\tref{tab:setting1} that when the difficulty level of the task

\subsection{How Challenging are Ref-Long Tasks?}
\label{sec:human}

As discussed in~\sref{sec:introduction}, a critical issue with existing long-context benchmarks is that they are often difficult for humans as well, thereby complicating efforts to assess their difficulty levels.
%while retrieval-based benchmarks are relatively easy for strong LCLMs. 
To this end, %following the findings in~\sref{sec:setting1 results} that Ref-Long tasks are challenging for LCLMs, 
%we conduct additional analyses as follows.
%\paragraph{Analysis I: Human Performance on Ref-Long tasks.} 
%As the extensive context length poses challenges for both models and human evaluation. We argue that Ref-Long can mitigate this issue and 
we conduct a human evaluation to see whether we can estimate the difficulty level of Ref-Long tasks through human performance. 

Specifically, we invite two annotators (PhD students) to complete 50 Ref-Long tasks that are randomly selected from the Multi-Hard-24K setting. For each task, annotators are provided with the same query used to prompt LCLMs. In addition, they are asked to complete each task at a pace that balances accuracy and efficiency, as the their time taken to solve each task is also recorded.
The annotators %achieve a Krippendorff’s alpha score~\cite{krippendorff2011computing} of XXXX, calculated by converting their results for each task into binary values—1 (correct) or 0 (incorrect), 
have the same and correct annotation on 42 tasks, indicating a high inter-agreement. Noting that %while humans may complete Ref-Long tasks within a reasonable time frame, LCLMs can solve these tasks within seconds, rendering direct comparisons unfair. Therefore, 
we also evaluate the o1~\cite{o1card} model, known for its strong reasoning abilities, %and longer reasoning time, 
for a more comprehensive comparison.

%\paragraph{Results.}
Experimental results of humans, GPT-4o and o1 on the 50 sampled tasks are listed in~\tref{tab:human performance}. As can be seen, while humans can correctly finish most of the Ref-Long tasks, the two LCLMs face significant challenges in addressing them. Due to the limited quota and poor performance of o1, we do not include it in the following experiments. Additionally, we calculate the average time required by annotators to complete a Ref-Long task as 123.95 seconds, which is reasonable considering the 24K input length.
%Notably, although GPT-4o completes Ref-Long tasks in a much shorter time, human annotators achieve far better performance at a speed comparable to that of o1. 
\textbf{These results demonstrate that Ref-Long tasks are manageable for humans but pose significant challenge for LCLMs, emphasizing the need for serious attention to these tasks in the development of LCLMs.} %\jw{this paragraph still needs some words regarding the time}

\begin{table}[tb]
\small
\centering
\setlength{\tabcolsep}{5mm}
\begin{tabular}{lcc}
\toprule
 & \textbf{F1}$\uparrow$ 
%& \textbf{$\text{P-Acc}_{\text{A}}$}$\uparrow$ & \textbf{$\text{P-Acc}_{\text{M}}$}$\uparrow$ 
& \textbf{Ex Acc}$\uparrow$ \\ %& \textbf{Time}\\
\midrule
GPT-4o & 74.24 & 14.00 \\%& $\sim$2.00 \\
o1 & 49.11 & 0.00 \\%& 33.64 \\
\midrule
Human & 99.08 & 92.00 \\%&123.95 \\
\bottomrule
\end{tabular}
\caption{Comparison of LCLMs and humans.} %\textbf{Time} is the average task finishing time, calculated in seconds.}
%\vspace{-0.2in}
\label{tab:human performance}
\end{table}

\subsection{Is Ref-Long Challenging Mainly Due to
the Instructions?}
\label{sec:factors}
%Given that humans can effectively solve Ref-Long tasks while LCLMs struggle, this section explores factors that may hinder LCLMs' performance on such tasks. We start by examine whether the issue arises from the format of the specific keys, which are represented as symbols rather than natural language. Moreover, %we split each document into smaller chunks to determine if the number of documents within a fixed context length influences performance. Finally, 
%we investigate whether this difficulty comes from LCLMs lacking an effective strategy to tackle Ref-Long tasks.

Given that humans can effectively solve Ref-Long tasks while LCLMs struggle, this section investigates whether the issue comes from LCLMs' unfamiliarity with the instructions used in~\sref{sec:setting1 data}. To this end, we incorporate a human-inspired strategy into the queries to provide hints for Ref-Long tasks, inspired by the chain-of-thought prompting approach~\cite{wei2022chain}.
Additionally, we examine whether the problem arises from the format of the specific keys in the instructions by converting the keys into natural language.

\paragraph{Incoporating Human Strategy.}
\label{sec:COT prompt}
\iffalse
\begin{table}[tb]
\renewcommand\arraystretch{1.1}
\centering
\setlength{\tabcolsep}{0.6mm}
\small
\begin{tabular}{lccc|ccc}
\toprule[1pt]
\multirow{3}*{\textbf{LCLM}} & \multicolumn{3}{c}{\textbf{Ex Acc}$\uparrow$} & \multicolumn{3}{c}{\textbf{F1}$\uparrow$} \\
\cmidrule(lr){2-4} \cmidrule(lr){5-7} 
 & 8K & 16K & 24K & 8K & 16K & 24K \\
\midrule[0.5pt]
LLama-3.1-Ins-8B &12.00 &2.00&0.00&60.14&45.97&38.85 \\
\qquad w/ Strategy &15.00&0.00&0.00&61.41&44.83&40.11 \\
LLama-3.1-Ins-70B & 61.00&13.00&4.00&87.38&63.87&52.21 \\
\qquad w/ Strategy  &34.00&5.00&1.00&69.64&49.29&39.96 \\
\midrule
GPT-4o-mini &71.00&18.00&7.00&92.10&79.54&68.64 \\
\qquad w/ Strategy &64.00&14.00&5.00&89.38&75.42&62.69 \\
GPT-4o &85.00 &43.00&19.00&94.81&86.61&75.38 \\
\qquad w/ Strategy &93.00&64.00&34.00&98.30&91.03&83.56 \\
\bottomrule[1pt]
\end{tabular}
\caption{Results with the human strategy-based prompts on the Multi-Hard setting of Ref-Long-A.}
\vspace{-0.2in}
\label{tab:hint results}
\end{table}
\fi

\begin{table}[tb]
\renewcommand\arraystretch{1.1}
\centering
\setlength{\tabcolsep}{0.6mm}
\small
\begin{tabular}{lccc|ccc}
\toprule[1pt]
\multirow{3}*{\textbf{LCLM}} & \multicolumn{3}{c}{\textbf{F1}$\uparrow$} & \multicolumn{3}{c}{\textbf{Ex Acc}$\uparrow$} \\
\cmidrule(lr){2-4} \cmidrule(lr){5-7} 
 & 8K & 16K & 24K & 8K & 16K & 24K \\
\midrule[0.5pt]
LLama-3.1-Ins-8B &60.14&45.97&38.85&12.00&2.00&0.00 \\
\qquad w/ Strategy &61.41&44.83&40.11&15.00&0.00&0.00 \\
LLama-3.1-Ins-70B &87.38&63.87&52.21&61.00&13.00&4.00 \\
\qquad w/ Strategy  &69.64&49.29&39.96&34.00&5.00&1.00 \\
\midrule
GPT-4o-mini &92.10&79.54&68.64&71.00&18.00&7.00 \\
\qquad w/ Strategy &89.38&75.42&62.69&64.00&14.00&5.00 \\
GPT-4o &94.81&86.61&75.38&85.00&43.00&19.00 \\
\qquad w/ Strategy &98.30&91.03&83.56&93.00&64.00&34.00 \\
\bottomrule[1pt]
\end{tabular}
\caption{Results with the human strategy-based prompts on the Multi-Hard setting of Ref-Long-A.}
%\vspace{-0.2in}
\label{tab:hint results}
\end{table}

Start by the large gap between LCLMs' and humans' performance on Ref-Long tasks, we investigate whether this difficulty comes from LCLMs' lack of an effective strategy to complete such tasks. When solving Ref-Long tasks, the two annotators in~\sref{sec:human} adopt a straightforward and effective method: dynamically constructing a dictionary while reading the input context. Specifically, when humans encounter a special key in the input for the first time, they add a new entry to the dictionary by pairing the key with the current document index. Otherwise, they simply update the existing entry by appending the corresponding document index to it. To teach LCLMs to use this human-like strategy, we explicitly include the above steps at the start of the prompts, and evaluate GPT and Llama-3-Ins models under the most challenging Multi-Hard-24K setting (See Appendix~\sref{appendix:prompts} for details of the prompt). 

\paragraph{Results.} 
The results are listed in~\tref{tab:hint results}. We observe that guiding GPT-4o with human strategy does enhance its performances, indicating that its referencing capability can be triggered with proper instructions. However, equipping weaker LCLMs with human strategies does not improve their performance, and even the enhanced results of GPT-4o are still far below human levels. These results suggest that improving the fundamental long-context referencing capabilities of LCLMs should take precedence over designing task-specific instructions or perform prompt engineering works. %if we aim to enhance LCLMs' performance on Ref-Long tasks.

\paragraph{Changing the Format of Keys.}
Next, we examine whether the format of specific keys in the instructions affects the ability of LCLMs to complete Ref-Long tasks. Specifically, we modify the Multi-Hard setting by replacing keys $\boldsymbol{num}$ $\in [0, 20)$ with 20 different fruit names, representing natural language. The prompts
are paraphrased accordingly to align with this modification (see Appendix~\sref{appendix:prompts} for the prompt template). We evaluate the GPT and Llama-3.1-Ins models in~\sref{sec:evaluated lclms} using these modified prompts, and list the results in~\tref{tab:symbol results}.

\paragraph{Results.}
As shown, switching the input keys to natural language does not lead to significant improvements and, in some cases, even lowers LCLMs' performance on Ref-Long tasks.  These results show that LCLMs are robust to variations in the format of specific keys when doing referencing tasks, and changing the format does not substantially impact the overall conclusions, which further supports us to construct the following Ref-Long subsets that include natural language keys. 
\iffalse
\begin{table}[tb]
\renewcommand\arraystretch{1.1}
\centering
\setlength{\tabcolsep}{0.6mm}
\small
\begin{tabular}{lccc|ccc}
\toprule[1pt]
\multirow{3}*{\textbf{LCLM}} & \multicolumn{3}{c}{\textbf{Ex Acc}$\uparrow$} & \multicolumn{3}{c}{\textbf{F1}$\uparrow$} \\
\cmidrule(lr){2-4} \cmidrule(lr){5-7} 
 & 8K & 16K & 24K & 8K & 16K & 24K \\
\midrule[0.5pt]
LLama-3.1-Ins-8B &12.00 &2.00&0.00&60.14&45.97&38.85 \\
\qquad \qquad w/ Fruit &15.00 &3.00&0.00&57.33&44.05&36.11 \\
LLama-3.1-Ins-70B & 61.00&13.00&4.00&87.38&63.87&52.21 \\
\qquad \qquad w/ Fruit  &63.00 &13.00&1.00&88.13&68.33&50.66  \\
\midrule
GPT-4o-mini & 71.00&18.00&7.00&92.10&79.54&68.64 \\
\qquad \qquad w/ Fruit &67.00 &20.00&3.00&90.63&79.34&68.24 \\
GPT-4o & 85.00&43.00&19.00&94.81&86.61&75.38 \\
\qquad \qquad w/ Fruit  &93.00&50.00&19.00&98.68&90.63&82.48 \\
\bottomrule[1pt]
\end{tabular}
\caption{Results with natural language keys on the Multi-Hard setting of Ref-Long-A.}
\vspace{-0.1in}
\label{tab:symbol results}
\end{table}
\fi

\begin{table}[tb]
\renewcommand\arraystretch{1.1}
\centering
\setlength{\tabcolsep}{0.6mm}
\small
\begin{tabular}{lccc|ccc}
\toprule[1pt]
\multirow{3}*{\textbf{LCLM}} & \multicolumn{3}{c}{\textbf{F1}$\uparrow$} & \multicolumn{3}{c}{\textbf{Ex Acc}$\uparrow$} \\
\cmidrule(lr){2-4} \cmidrule(lr){5-7} 
 & 8K & 16K & 24K & 8K & 16K & 24K \\
\midrule[0.5pt]
LLama-3.1-Ins-8B &60.14&45.97&38.85&12.00 &2.00&0.00 \\
\qquad \qquad w/ Fruit &57.33&44.05&36.11&15.00 &3.00&0.00 \\
LLama-3.1-Ins-70B &87.38&63.87&52.21&61.00&13.00&4.00 \\
\qquad \qquad w/ Fruit  &88.13&68.33&50.66&63.00 &13.00&1.00 \\
\midrule
GPT-4o-mini &92.10&79.54&68.64&71.00&18.00&7.00 \\
\qquad \qquad w/ Fruit &90.63&79.34&68.24&67.00 &20.00&3.00 \\
GPT-4o &94.81&86.61&75.38&85.00&43.00&19.00 \\
\qquad \qquad w/ Fruit  &98.68&90.63&82.48&93.00&50.00&19.00 \\
\bottomrule[1pt]
\end{tabular}
\caption{Results with natural language keys on the Multi-Hard setting of Ref-Long-A.}
%\vspace{-0.1in}
\label{tab:symbol results}
\end{table}

\iffalse
\paragraph{Factor III: Length of Individual Document.}
\label{sec:shorter document}
Finally, we investigate whether the length of the $\boldsymbol{M}$ documents in each task affects LCLMs' performance. %, inspired by a hypothesis that shorter documents may enable models to associate a specific key with its surrounding context easier. 
Specifically, we truncate these documents under the multi-hard setting into several sub-documents of equal length and re-index these sub-documents accordingly. We then evaluate GPT and Llama models on these newly-generated tasks and show the results in Figure XX.

\begin{table}[tb]
\renewcommand\arraystretch{1.1}
\centering
\setlength{\tabcolsep}{1mm}
\small
\begin{tabular}{lccc|ccc}
\toprule[1pt]
\multirow{2}*{LCLM} & \multicolumn{3}{c}{\textbf{Acc}$\uparrow$} & \multicolumn{3}{c}{\textbf{F1}$\uparrow$} \\
 & 8K & 16K & 24K & 8K & 16K & 24K \\
\midrule[0.5pt]
Ori & 85.00&33.00&11.00&94.92&74.66 &55.39\\
truncate*2  &90.00&55.00&18.00&96.75&85.90&63.98 \\
truncate*4  &97.00&49.00&23.00&99.22&82.40&65.60 \\
truncate*8   &98.00&59.00&26.00&99.25&85.42&71.79 \\
\bottomrule[1pt]
\end{tabular}
\caption{GPT-4o results with different chunk size.}
%\vspace{-0.2in}
\label{tab:chunk results}
\end{table}
\paragraph{Results.} 
aa
\fi
\textbf{Overall, this section illustrates the challenges LCLMs are facing on Ref-Long tasks cannot be addressed by adjusting superficial factors during inference.}

\subsection{Could Fine-tuning Solve Ref-Long?}
\label{sec:fine-tuning}

\begin{table}[tb]
\renewcommand\arraystretch{1.1}
\centering
\setlength{\tabcolsep}{0.6mm}
\small
\begin{tabular}{lccc|ccc}
\toprule[1pt]
\multirow{3}*{\textbf{LCLM}} & \multicolumn{3}{c}{\textbf{Single}} & \multicolumn{3}{c}{\textbf{Multi}} \\
\cmidrule(lr){2-4} \cmidrule(lr){5-7} 
 & 8K & 16K & 24K & 8K & 16K & 24K \\
\midrule[0.5pt]
LLama-3.1-Ins-8B &59.00&20.00&13.00&17.00&5.00&2.00 \\
\quad w/ FT-Multi-Easy &50.00&47.00&45.00&22.00&20.00&8.00 \\
\bottomrule[1pt]
\end{tabular}
\caption{Ex Acc scores of LCLMs on the Easy setting of Ref-Long-A.}
%\vspace{-0.2in}
\label{tab:ft results}
\end{table}

\begin{table*}[tb]
\small
\centering
\setlength{\tabcolsep}{1mm}
\begin{tabular}{l  *{7}{c} | l  *{7}{c}}
\toprule
  \multirow{3}{*}{\textbf{LCLM}} &
  \multicolumn{7}{c}{\textbf{F1}$\uparrow$} & \multicolumn{7}{c}{\textbf{Ex Acc}$\uparrow$} \\
\cmidrule(lr){2-8} \cmidrule(lr){9-15} 
 & 8K & 16K & 24K & 32K & 40K & 48K & 56K & 8K & 16K & 24K & 32K & 40K & 48K & 56K \\
\midrule
 LongCite-8B &20.90&10.47&12.45&7.92&9.99&7.26&7.80&4.00&0.00&0.00&0.00&0.00&0.00&0.00 \\
 Qwen2.5-Ins-7B &49.98&38.81&39.21&33.59&29.77&27.04&24.87&16.00&1.00&0.00&0.00&0.00&0.00&0.00 \\
 Phi-3-mini &53.72&37.84&33.54&28.42&26.11&23.92&22.92&26.00&4.00&0.00&0.00&0.00&0.00&0.00 \\
 ProLong-8B-64K &70.02&52.36&45.66&36.83&30.11&26.80&24.06&33.00&5.00&3.00&0.00&0.00&0.00&0.00 \\
 ProLong-8B-512K &73.84&52.39&42.75&37.40&31.10&29.42&25.95&39.00&6.00&1.00&0.00&0.00&0.00&0.00 \\
 Llama-3.1-Ins-8B &81.86&63.05&52.37&41.70&37.19&35.21&30.53&52.00&14.00&7.00&2.00&2.00&0.00&0.00 \\
 Qwen2.5-Ins-72B &83.43&76.35&74.92&71.30&68.00&64.87&62.00&55.00&34.00&19.00&10.00&6.00&0.00&1.00 \\
 GPT-4o mini &89.85&80.53&76.67&71.30&63.48&60.73&55.71&70.00&39.00&22.00&15.00&4.00&3.00&1.00 \\
 Llama-3.1-Ins-70B &\textbf{91.87}&\textbf{86.21}&80.08&72.01&61.43&54.35&48.25&77.00&54.00&34.00&18.00&8.00&4.00&4.00 \\
 Gemini-1.5-Flash &89.97&86.05&79.45&76.47&74.27&72.74&70.85&77.00&52.00&27.00&15.00&11.00&10.00&6.00 \\
 Gemini-1.5-Pro &90.10&84.50&80.47&76.74&68.33&62.80&61.20&\textbf{78.00}&\textbf{56.00}&39.00&28.00&15.00&\textbf{12.00}&\textbf{10.00} \\
 GPT-4o &90.00&84.93&\textbf{83.50}&\textbf{80.85}&\textbf{80.43}&\textbf{75.96}&\textbf{73.81}&71.00&\textbf{56.00}&\textbf{41.00}&\textbf{32.00}&\textbf{20.00}&11.00&8.00 \\
\bottomrule
\end{tabular}
\caption{Results on Ref-Long-F. 
The best results under each column are \textbf{boldfaced}. For conciseness, we only list the results of the Twitter topic here, and show the results of the rest two topics in~\tref{tab:complete setting1}.
}
%\vspace{-0.2in}
\label{tab:setting2}
\end{table*}

%\jw{The transition here can be modified.}
Existing research has demonstrated that fine-tuning LCLMs using (1) carefully designed strategies~\cite{gao2024train, bai2024longalign} and (2) QA pairs with sentence-level references~\cite{zhang2024longcite} enhances their long-context understanding capabilities. To explore this claim in the context of long-context referencing, we also evaluate fine-tuned models from these two categories—Prolong-8B-64K/512K for (1) and LongCite-8B for (2)—and present the results in ~\tref{tab:setting1}. We observe that ProLong-8B-512K achieves results comparable to Llama-3.1-Ins-8B while requiring significantly less data and being fine-tuned on a weaker backbone (Llama-3-Ins-8B)~\cite{gao2024train}. This highlights the importance of a well-designed long-context fine-tuning strategy. In contrast, LongCite-8B underperforms its backbone, Llama-3.1-Ins-8B, showing that models fine-tuned on specific long-context tasks may not generalize well on long-context referencing tasks.

Additionally, we study whether fine-tuning on Ref-Long tasks can enhance LCLM performance. Due to computational constraints, we adopt the following steps: (1) Fine-tune Llama-3.1-Ins-8B, and (2) use the Multi-Easy-8K setting to construct the fine-tuning data, as the Medium/Hard settings are too challenging for Llama-3.1-Ins-8B, and fine-tuning on 16K/24K data exceeds our computational capacity. Specifically, we create 500 Multi-Easy 8K tasks that do not overlap with those in~\tref{tab:setting1}, and fine-tune Llama-3.1-Ins-8B on this data (check Appendix~\sref{appendix:inference details} for fine-tuning details). The fine-tuned model is then evaluated on the Easy setting, with results shown in~\tref{tab:ft results}. We find that fine-tuning on Ref-Long tasks does boost model performance, particularly on easier tasks. However, as input length and task difficulty increase, all the above fine-tuned models still perform poorly, \textbf{demonstrating that fine-tuning on long-context data alone cannot fully overcome LCLMs' limitations on Ref-Long tasks}.

\section{Extending to Realistic Scenarios}
\label{sec:realistic}

\subsection{Fluent Key in Fluent Context}
\label{sec:fluent with fluent}

\paragraph{Dataset.}
\label{sec:setting2 data}

In~\sref{sec:factors}, we observe that changing the format of specific keys to natural language does not significantly affect LCLMs' performance on Ref-Long tasks, which motivates us to extend Ref-Long to a more realistic scenario, where keys are spans of text embedded within a coherent context (e.g., legal terms within law books).
%, which contrasts with the dataset in~\sref{sec:setting1 data}, where keys are not contextually coherent with their surroundings. 
%To explore whether the conclusions drawn from the first dataset apply to this more realistic scenario,
To this end, we construct the second subset \textbf{Ref-Long-F}luent (\textbf{Ref-Long-F}) for investigation.   

Specifically, this subset is constructed upon the SummHay benchmark~\cite{laban2024summary}, which comprises 10 topics, each associated with 100 documents. For each topic, there is a candidate set of insights—short statements containing specific information about the topic. Each document is generated by sampling 3–8 distinct insights from the candidate set and using GPT-4o to create a coherent 1,000-token document incorporating all the sampled insights (slight paraphrasing insights is allowed), which makes such insights suitable for acting as specific keys. When constructing the subset, we select 3 news topics (Foot Locker, Twitter, Financial Market) from SummHay. Under each topic, we set the number of sampled documents $\boldsymbol{M} \in \{8,16,24,32,40,48,56\}$ (corresponding to input lengths ranging from 8K to 56K tokens) and create 100 tasks for each length with the aggregative method mentioned in~\sref{sec:setting1 data}, resulting in a subset of 2,100 tasks. We then evaluate all the LCLMs on these tasks, with the results presented in~\tref{tab:setting2} (see Appendix~\sref{appendix:prompts} for prompt details). Following the \textit{fluent key in fluent context} setting, we additionally construct a real-world dataset for completeness and evaluate all models on it in Appendix~\ref{appendix:nba data}.

\paragraph{Results.}
\label{sec:setting2 results}
We find that LCLMs' performances on Ref-Long-F align with results on Ref-Long-A in that larger models consistently outperform smaller ones across different input lengths, although the rankings of individual LCLMs may vary. For instance, Gemini-1.5-Pro achieves results comparable to GPT-4o on Ref-Long-F but lags behind on Ref-Long-A, reflecting its stronger capability in referencing natural language keys. Furthermore, all LCLMs exhibit decreasing performance as input length increases, reaffirming the findings in~\fref{fig:ablation}. Notably, none of the LCLMs achieve Ex Acc scores above 20.00\% when input lengths reach 40K tokens, falling largely short of their claimed maximum input capabilities. \textbf{Overall, we conclude that LCLMs' lack of referencing capability occurs on both incoherent and coherent documents, further emphasizing the severity of this issue.}

%Moreover, similar to~\tref{tab:setting1}, all LCLMs have much larger F1 scores compared to Ex Acc scores. 
%Moreover, similar to~\tref{tab:setting1}, all LCLMs have much larger F1 scores compared to Ex Acc scores, and we conduct an error analysis in~\sref{sec:error2} to further study this phenomenon.

\subsection{Paper Citation}
\label{sec:paper citation}

\begin{table*}[tb]
\small
\centering
\setlength{\tabcolsep}{2.3mm}
\begin{tabular}{l *{4}{c} | *4{c}}
\toprule
  \multirow{3}{*}{\textbf{LCLM}} 
 %& \multicolumn{8}{c}{\textbf{Easy}} \\
 & \multicolumn{4}{c}{\textbf{F1}$\uparrow$} & \multicolumn{4}{c}{\textbf{Ex Acc}$\uparrow$} \\
  \cmidrule(lr){2-5} \cmidrule(lr){6-9}
 & 8 (30K) & 12 (45K) & 16 (60K)& 20 (75K)
 & 8 (30K) & 12 (45K) & 16 (60K)& 20 (75K)\\
\midrule
LongCite-8B  &0.40  &0.46  &0.80  &1.00  &0.00  &0.00  &0.00  &0.00  \\
Phi-3-mini  &11.36  &8.78  & 9.24 & 5.23  & 1.00 & 0.00 &0.00  &0.00  \\
ProLong-8B-64K  & 11.89 & 12.32 & 12.07 & - & 0.00 & 0.00 & 0.00 & - \\
Qwen2.5-Ins-7B  &24.29  &19.50  &19.80  &10.37  & 0.00 &0.00  &0.00  &0.00  \\
Llama-3.1-Ins-8B  & 21.46 & 30.17 & 27.63 & 18.54  & 2.00 & 2.00 & 0.00 &1.00  \\
ProLong-8B-512K  & 15.49 & 12.78 & 14.20 & 13.34  & 0.00 & 0.00 & 0.00 &0.00  \\
Llama-3.1-Ins-70B  & 64.39 & 53.88 & 46.65 & 30.28  & 26.00 & 6.00 & 5.00 & 2.00 \\
GPT-4o mini  &81.30  &78.10  &64.02  &60.12  &45.00 &34.00  &22.00  &14.00  \\
Gemini-1.5-Flash  & 80.42 &77.44  &78.39  &75.41  & 34.00 &29.00  &32.00  &25.00  \\
Gemini-1.5-Pro  &84.83  &78.63  &70.89  &70.89  & 52.00 &41.00  &29.00  &20.00  \\
Qwen2.5-Ins-72B  &86.59  &79.65  &\textbf{80.48}  &\textbf{76.69}  & 55.00 &42.00  &\textbf{40.00}  &\textbf{31.00}  \\
GPT-4o  &\textbf{91.81}  &\textbf{87.30}  &78.22  &71.13  &\textbf{66.00} &\textbf{48.00}  &30.00  &17.00  \\
\bottomrule
\end{tabular}
\caption{Results on Ref-Long-Paper. 
The best results under each column are \textbf{boldfaced}. Since papers are not 1000 tokens long, we only list the number of $\boldsymbol{M}$ and provide the average input length of tasks under each $\boldsymbol{M}$ in brackets. ``-'' means the input length exceeds the maximum input length of a model.}
%\vspace{-0.2in}
\label{tab:setting3}
\end{table*}

\paragraph{Dataset.}
As mentioned in~\sref{sec:introduction}, long-context referencing tasks play a crucial role in many real-world applications. Therefore, having evaluated LCLMs on two synthetic subsets, we now turn our attention to their performances on Ref-Long tasks constructed from real-world data. Due to the lack of appropriate datasets for this purpose, we manually construct the third subset targeting citations of computer science arXiv papers (\textbf{Ref-Long-Paper}). 

Specifically, we first collected 47 arXiv papers that meet the following criteria: (1) published after March 2024; (2) cited by at least two papers that shorter than 5,000 tokens; as our seed papers. Additionally, we collected 34 extra arXiv papers published between January and February 2024 that are shorter than 5000 tokens to serve as distractors, since these papers cannot cite the seed papers due to the publishing time. Note that we control the lengths of papers since they are much longer than the documents used in the previous subsets, and 5000 tokens is actually short for papers. Also, we select papers published after 2024 to avoid overlap with the training data of most evaluated LCLMs. 

To create an Ref-Long task with \(\boldsymbol{M}\) papers, we first randomly select one seed paper, then use its \(\boldsymbol{m}\) citations and \(\boldsymbol{M} - \boldsymbol{m}\) randomly sampled distractors to form the task input. Using this approach, we create 100 tasks for each \(\boldsymbol{M} \in \{8, 12, 16, 20\}\). The aggregative sampling method described in~\sref{sec:setting1 data} is also applied as \(\boldsymbol{M}\) increases. During evaluation, LCLMs are tasked with identifying the indexes of papers citing the seed paper, using its title as the key (see prompts in Appendix~\sref{appendix:prompts}). The maximum value of \(\boldsymbol{m}\) is 6 in our data, ensuring that distractors are included in every task.

\paragraph{Results.}
\label{sec:setting3 results}
We present the results on Ref-Long-Paper in~\tref{tab:setting3}. As shown, smaller LCLMs struggle to complete the paper referencing tasks, likely due to the larger token counts in research papers compared to the documents in the previous two subsets. Even for the stronger LCLMs, their performances are unsatisfactory, with around 30\% Ex Acc scores for tasks with only 16 papers. An exception is Qwen2.5-Ins-72B, which even outperforms Gemini-1.5-Pro. We hypothesize this may be because some of the arXiv papers used in~\tref{tab:setting3} are part of Qwen2.5-Ins-72B's pre-training data, given its release date of September 2024. This result again supports Ref-Long's design of including both synthetic and real-world subsets, where LCLMs' results on synthetic subsets are more aligning with their sizes since data contamination is avoided. 

%even the strongest LCLMs still struggle with the paper citation tasks with only 20 candidate papers, which aligns with our observations on the previous two datasets.

\section{Error Analysis.}
\label{sec:error}

\begin{table}[tb]
\small
\centering
\setlength{\tabcolsep}{2mm}
\begin{tabular}{l|c|ccc}
\toprule
\textbf{Subset}& \textbf{Num}&\textbf{Type I}& \textbf{Type II} & \textbf{Type III} %\textbf{Type IV} 
\\
\midrule[0.5pt]
Ref-Long-A &81&85.19&1.23&13.58 \\
Ref-Long-F &59&37.29&50.85&11.86 \\
Ref-Long-Paper &83&12.05&54.22&33.73\\
\bottomrule
\end{tabular}
\caption{Percentage (\%) of  GPT-4o's error types on the Multi-Hard-24K, 24K, and 20 settings of the three subsets. ``Num'' refers to the number of errors.}
%\vspace{-0.2in}
\label{tab:error}
\end{table}

Given LCLMs' poor performances on the three Ref-Long subsets, %~\tref{tab:setting1}, \tref{tab:setting2}, and~\tref{tab:setting3}, 
we further investigate what types of errors they tend to produce on these three subsets. Specifically, we first manually go through $\sim$100 errors of LCLMs across these subsets and conclude three general error types: \textit{reference less} (I), \textit{reference more} (II), and \textit{both} (III). We annotate GPT-4o's failure cases on the same input length--Multi-Hard-24K setting of Ref-Long-A, 24K setting of Ref-Long-F, and the 20 setting of Ref-Long-Paper, and list the percentage of each error type in~\tref{tab:error}. 

We observe that when the specific key consists of a number along with a specific symbol (Ref-Long-A), GPT-4o rarely confuses the key with other keys. However, it is not sensitive enough to the specific key, with most errors resulting from failing to identify all documents referencing the key (type I errors). Conversely, when the specific key is formatted in natural language in the other two subsets, GPT-4o often confuses this key with others and includes documents only referencing unrelated keys in its answers. These results indicate that as context length increases, LCLMs may become either overly sensitive or insufficiently sensitive to keys in the given documents, both of which contribute to their poor performances on Ref-Long tasks.

\section{Conclusion}
This paper addresses limitations in existing long-context benchmarks by introducing Ref-Long, a novel benchmark designed to systematically evaluate the long-context referencing capability of LCLMs. Ref-Long requires LCLMs to generate indexes of documents referencing a specific key, a task that proves to be difficult even for the most advanced LCLMs at input lengths far shorter than their claimed maximum context sizes. Based on these observations, we conduct extensive investigations and show that neither adjusting query format nor fine-tuning sufficiently solves Ref-Long tasks. Finally, we extend Ref-Long to more realistic scenarios, revealing that limitations in LCLMs' referencing capabilities persist, and encourage future works to enhance real-world long-context systems by focusing on the referencing capability.

\section*{Limitations}
Due to budget constraints, we evaluate only Llama-3.3-Ins-70B in~\tref{tab:setting1} and Gemini-1.5-Pro in the experiments presented in the main paper. While this may introduce potential bias in the evaluation results, we argue that the existing findings are sufficiently convincing. We plan to provide more comprehensive results in future work when additional experimental resources become available.

Additionally, for Ref-Long-F, which is based on the SummHay benchmark, it is important to note that SummHay covers only a limited set of topics. This limitation implies that the evaluation may not fully capture the performance of LCLMs across a broader range of topics. As a result, the effectiveness of Ref-Long in scenarios involving topics beyond those covered by SummHay remains uncertain. Future work could address this by expanding the topic diversity within the benchmark, enabling a more comprehensive evaluation of LCLMs' referencing capabilities across various topics.

\section*{Ethical Considerations}
Since this paper includes responses generated by LCLMs, it is possible that these model-generated contents may contain toxic or harmful information, necessitating comprehensive data processing by users. Additionally, one of our subsets (Ref-Long-F) is based on an existing benchmark generated by LLMs, which may also include toxic or harmful information. Although we have manually reviewed the data, it is still possible that the original benchmark contains such content, requiring further processing.

\section*{Acknowledgment}
This work has been made possible by a Research Impact Fund project (RIF R6003-21) and a General Research Fund project (GRF 16203224) funded by the Research Grants Council (RGC) of the Hong Kong Government.
We are also grateful for the TPU compute support provided by the Google TRC program.

%\input{acl_latex.bbl} 
\bibliography{custom}

\begin{thebibliography}{41}
\providecommand{\natexlab}[1]{#1}

\bibitem[{Abdin et~al.(2024)Abdin, Jacobs, Awan, Aneja, Awadallah, Awadalla, Bach, Bahree, Bakhtiari, Behl, Benhaim, Bilenko, Bjorck, Bubeck, Cai, Mendes, Chen, Chaudhary, Chopra, Giorno, de~Rosa, Dixon, Eldan, Iter, Garg, Goswami, Gunasekar, Haider, Hao, Hewett, Huynh, Javaheripi, Jin, Kauffmann, Karampatziakis, Kim, Khademi, Kurilenko, Lee, Lee, Li, Liang, Liu, Lin, Lin, Madan, Mitra, Modi, Nguyen, Norick, Patra, Perez-Becker, Portet, Pryzant, Qin, Radmilac, Rosset, Roy, Ruwase, Saarikivi, Saied, Salim, Santacroce, Shah, Shang, Sharma, Song, Tanaka, Wang, Ward, Wang, Witte, Wyatt, Xu, Xu, Yadav, Yang, Yang, Yu, Zhang, Zhang, Zhang, Zhang, Zhang, Zhang, Zhang, and Zhou}]{abdin2024phi3}
Marah Abdin, Sam~Ade Jacobs, Ammar~Ahmad Awan, Jyoti Aneja, Ahmed Awadallah, Hany Awadalla, Nguyen Bach, Amit Bahree, Arash Bakhtiari, Harkirat Behl, Alon Benhaim, Misha Bilenko, Johan Bjorck, Sébastien Bubeck, Martin Cai, Caio César~Teodoro Mendes, Weizhu Chen, Vishrav Chaudhary, Parul Chopra, Allie~Del Giorno, Gustavo de~Rosa, Matthew Dixon, Ronen Eldan, Dan Iter, Amit Garg, Abhishek Goswami, Suriya Gunasekar, Emman Haider, Junheng Hao, Russell~J. Hewett, Jamie Huynh, Mojan Javaheripi, Xin Jin, Piero Kauffmann, Nikos Karampatziakis, Dongwoo Kim, Mahoud Khademi, Lev Kurilenko, James~R. Lee, Yin~Tat Lee, Yuanzhi Li, Chen Liang, Weishung Liu, Eric Lin, Zeqi Lin, Piyush Madan, Arindam Mitra, Hardik Modi, Anh Nguyen, Brandon Norick, Barun Patra, Daniel Perez-Becker, Thomas Portet, Reid Pryzant, Heyang Qin, Marko Radmilac, Corby Rosset, Sambudha Roy, Olatunji Ruwase, Olli Saarikivi, Amin Saied, Adil Salim, Michael Santacroce, Shital Shah, Ning Shang, Hiteshi Sharma, Xia Song, Masahiro Tanaka, Xin Wang, Rachel
  Ward, Guanhua Wang, Philipp Witte, Michael Wyatt, Can Xu, Jiahang Xu, Sonali Yadav, Fan Yang, Ziyi Yang, Donghan Yu, Chengruidong Zhang, Cyril Zhang, Jianwen Zhang, Li~Lyna Zhang, Yi~Zhang, Yue Zhang, Yunan Zhang, and Xiren Zhou. 2024.
\newblock \href {https://arxiv.org/abs/2404.14219} {Phi-3 technical report: A highly capable language model locally on your phone}.
\newblock \emph{Preprint}, arXiv:2404.14219.

\bibitem[{AI@Meta(2024)}]{llama3.1modelcard}
AI@Meta. 2024.
\newblock \href {https://ai.meta.com/blog/meta-llama-3-1/} {Introducing llama 3.1: Our most capable models to date}.

\bibitem[{An et~al.(2024)An, Gong, Zhong, Zhao, Li, Zhang, Kong, and Qiu}]{an-etal-2024-l}
Chenxin An, Shansan Gong, Ming Zhong, Xingjian Zhao, Mukai Li, Jun Zhang, Lingpeng Kong, and Xipeng Qiu. 2024.
\newblock \href {https://aclanthology.org/2024.acl-long.776} {{L}-eval: Instituting standardized evaluation for long context language models}.
\newblock In \emph{Proceedings of the 62nd Annual Meeting of the Association for Computational Linguistics (Volume 1: Long Papers)}, pages 14388--14411, Bangkok, Thailand. Association for Computational Linguistics.

\bibitem[{Bai et~al.(2024{\natexlab{a}})Bai, Lv, Gu, Liu, Zou, Cao, Hou, Dong, Feng, Li et~al.}]{bai2024longcite}
Yushi Bai, Xin Lv, Wanjun Gu, Danqing Liu, Minhao Zou, Shulin Cao, Lei Hou, Yuxiao Dong, Ling Feng, Juanzi Li, et~al. 2024{\natexlab{a}}.
\newblock Longcite: Enabling llms to generate fine-grained citations in long-context qa.
\newblock \emph{arXiv preprint arXiv:2409.02897}.

\bibitem[{Bai et~al.(2024{\natexlab{b}})Bai, Lv, Zhang, He, Qi, Hou, Tang, Dong, and Li}]{bai2024longalign}
Yushi Bai, Xin Lv, Jiajie Zhang, Yuze He, Ji~Qi, Lei Hou, Jie Tang, Yuxiao Dong, and Juanzi Li. 2024{\natexlab{b}}.
\newblock Longalign: A recipe for long context alignment of large language models.
\newblock In \emph{Findings of the Association for Computational Linguistics: EMNLP 2024}, pages 1376--1395.

\bibitem[{Bai et~al.(2024{\natexlab{c}})Bai, Lv, Zhang, Lyu, Tang, Huang, Du, Liu, Zeng, Hou et~al.}]{bai2024longbench}
Yushi Bai, Xin Lv, Jiajie Zhang, Hongchang Lyu, Jiankai Tang, Zhidian Huang, Zhengxiao Du, Xiao Liu, Aohan Zeng, Lei Hou, et~al. 2024{\natexlab{c}}.
\newblock Longbench: A bilingual, multitask benchmark for long context understanding.
\newblock In \emph{Proceedings of the 62nd Annual Meeting of the Association for Computational Linguistics (Volume 1: Long Papers)}, pages 3119--3137.

\bibitem[{Bao et~al.(2021)Bao, Wu, Zhang, Chandrasekharan, and Jurgens}]{bao2021conversations}
Jiajun Bao, Junjie Wu, Yiming Zhang, Eshwar Chandrasekharan, and David Jurgens. 2021.
\newblock Conversations gone alright: Quantifying and predicting prosocial outcomes in online conversations.
\newblock In \emph{Proceedings of the Web Conference 2021}, pages 1134--1145.

\bibitem[{Gao et~al.(2024)Gao, Wettig, Yen, and Chen}]{gao2024train}
Tianyu Gao, Alexander Wettig, Howard Yen, and Danqi Chen. 2024.
\newblock How to train long-context language models (effectively).
\newblock \emph{arXiv preprint arXiv:2410.02660}.

\bibitem[{Gao et~al.(2023)Gao, Yen, Yu, and Chen}]{gao2023enabling}
Tianyu Gao, Howard Yen, Jiatong Yu, and Danqi Chen. 2023.
\newblock Enabling large language models to generate text with citations.
\newblock In \emph{Proceedings of the 2023 Conference on Empirical Methods in Natural Language Processing}, pages 6465--6488.

\bibitem[{Gemini(2024)}]{geminiteam2024gemini}
Gemini. 2024.
\newblock \href {https://arxiv.org/abs/2403.05530} {Gemini 1.5: Unlocking multimodal understanding across millions of tokens of context}.
\newblock \emph{Preprint}, arXiv:2403.05530.

\bibitem[{Hsieh et~al.(2024)Hsieh, Sun, Kriman, Acharya, Rekesh, Jia, and Ginsburg}]{hsieh2024ruler}
Cheng-Ping Hsieh, Simeng Sun, Samuel Kriman, Shantanu Acharya, Dima Rekesh, Fei Jia, and Boris Ginsburg. 2024.
\newblock Ruler: What's the real context size of your long-context language models?
\newblock \emph{arXiv preprint arXiv:2404.06654}.

\bibitem[{Hu et~al.()Hu, Wallis, Allen-Zhu, Li, Wang, Wang, Chen et~al.}]{hulora}
Edward~J Hu, Phillip Wallis, Zeyuan Allen-Zhu, Yuanzhi Li, Shean Wang, Lu~Wang, Weizhu Chen, et~al.
\newblock Lora: Low-rank adaptation of large language models.
\newblock In \emph{International Conference on Learning Representations}.

\bibitem[{Hurst et~al.(2024)Hurst, Lerer, Goucher, Perelman, Ramesh, Clark, Ostrow, Welihinda, Hayes, Radford, Madry, Baker{-}Whitcomb, Beutel, Borzunov, Carney, Chow, Kirillov, Nichol, Paino, Renzin, Passos, Kirillov, Christakis, Conneau, Kamali, Jabri, Moyer, Tam, Crookes, Tootoonchian, Kumar, Vallone, Karpathy, Braunstein, Cann, Codispoti, Galu, Kondrich, Tulloch, Mishchenko, Baek, Jiang, Pelisse, Woodford, Gosalia, Dhar, Pantuliano, Nayak, Oliver, Zoph, Ghorbani, Leimberger, Rossen, Sokolowsky, Wang, Zweig, Hoover, Samic, McGrew, Spero, Giertler, Cheng, Lightcap, Walkin, Quinn, Guarraci, Hsu, Kellogg, Eastman, Lugaresi, Wainwright, Bassin, Hudson, Chu, Nelson, Li, Shern, Conger, Barette, Voss, Ding, Lu, Zhang, Beaumont, Hallacy, Koch, Gibson, Kim, Choi, McLeavey, Hesse, Fischer, Winter, Czarnecki, Jarvis, Wei, Koumouzelis, and Sherburn}]{gpt4o}
Aaron Hurst, Adam Lerer, Adam~P. Goucher, Adam Perelman, Aditya Ramesh, Aidan Clark, AJ~Ostrow, Akila Welihinda, Alan Hayes, Alec Radford, Aleksander Madry, Alex Baker{-}Whitcomb, Alex Beutel, Alex Borzunov, Alex Carney, Alex Chow, Alex Kirillov, Alex Nichol, Alex Paino, Alex Renzin, Alex~Tachard Passos, Alexander Kirillov, Alexi Christakis, Alexis Conneau, Ali Kamali, Allan Jabri, Allison Moyer, Allison Tam, Amadou Crookes, Amin Tootoonchian, Ananya Kumar, Andrea Vallone, Andrej Karpathy, Andrew Braunstein, Andrew Cann, Andrew Codispoti, Andrew Galu, Andrew Kondrich, Andrew Tulloch, Andrey Mishchenko, Angela Baek, Angela Jiang, Antoine Pelisse, Antonia Woodford, Anuj Gosalia, Arka Dhar, Ashley Pantuliano, Avi Nayak, Avital Oliver, Barret Zoph, Behrooz Ghorbani, Ben Leimberger, Ben Rossen, Ben Sokolowsky, Ben Wang, Benjamin Zweig, Beth Hoover, Blake Samic, Bob McGrew, Bobby Spero, Bogo Giertler, Bowen Cheng, Brad Lightcap, Brandon Walkin, Brendan Quinn, Brian Guarraci, Brian Hsu, Bright Kellogg, Brydon
  Eastman, Camillo Lugaresi, Carroll~L. Wainwright, Cary Bassin, Cary Hudson, Casey Chu, Chad Nelson, Chak Li, Chan~Jun Shern, Channing Conger, Charlotte Barette, Chelsea Voss, Chen Ding, Cheng Lu, Chong Zhang, Chris Beaumont, Chris Hallacy, Chris Koch, Christian Gibson, Christina Kim, Christine Choi, Christine McLeavey, Christopher Hesse, Claudia Fischer, Clemens Winter, Coley Czarnecki, Colin Jarvis, Colin Wei, Constantin Koumouzelis, and Dane Sherburn. 2024.
\newblock Gpt-4o system card.
\newblock \emph{CoRR}, abs/2410.21276.

\bibitem[{Kamradt(2023{\natexlab{a}})}]{needle}
Greg Kamradt. 2023{\natexlab{a}}.
\newblock \href {https://github.com/gkamradt/LLMTest_NeedleInAHaystack} {Needle in a haystack - pressure testing llms}.

\bibitem[{Kamradt(2023{\natexlab{b}})}]{kamradt2023needle}
Gregory Kamradt. 2023{\natexlab{b}}.
\newblock Needle in a haystack - pressure testing llms.
\newblock \url{https://github.com/gkamradt/LLMTest_NeedleInAHaystack/tree/main}.

\bibitem[{Karpinska et~al.(2024)Karpinska, Thai, Lo, Goyal, and Iyyer}]{karpinska2024one}
Marzena Karpinska, Katherine Thai, Kyle Lo, Tanya Goyal, and Mohit Iyyer. 2024.
\newblock One thousand and one pairs: A “novel” challenge for long-context language models.
\newblock In \emph{Proceedings of the 2024 Conference on Empirical Methods in Natural Language Processing}, pages 17048--17085.

\bibitem[{Kuratov et~al.(2024)Kuratov, Bulatov, Anokhin, Rodkin, Sorokin, Sorokin, and Burtsev}]{kuratov2024babilong}
Yury Kuratov, Aydar Bulatov, Petr Anokhin, Ivan Rodkin, Dmitry Sorokin, Artyom Sorokin, and Mikhail Burtsev. 2024.
\newblock Babilong: Testing the limits of llms with long context reasoning-in-a-haystack.
\newblock \emph{Advances in Neural Information Processing Systems}, 37:106519--106554.

\bibitem[{Kwon et~al.(2023)Kwon, Li, Zhuang, Sheng, Zheng, Yu, Gonzalez, Zhang, and Stoica}]{kwon2023efficient}
Woosuk Kwon, Zhuohan Li, Siyuan Zhuang, Ying Sheng, Lianmin Zheng, Cody~Hao Yu, Joseph~E. Gonzalez, Hao Zhang, and Ion Stoica. 2023.
\newblock Efficient memory management for large language model serving with pagedattention.
\newblock In \emph{Proceedings of the ACM SIGOPS 29th Symposium on Operating Systems Principles}.

\bibitem[{Laban et~al.(2024)Laban, Fabbri, Xiong, and Wu}]{laban2024summary}
Philippe Laban, Alexander~Richard Fabbri, Caiming Xiong, and Chien-Sheng Wu. 2024.
\newblock Summary of a haystack: A challenge to long-context llms and rag systems.
\newblock In \emph{Proceedings of the 2024 Conference on Empirical Methods in Natural Language Processing}, pages 9885--9903.

\bibitem[{Levy et~al.(2024)Levy, Jacoby, and Goldberg}]{levy2024same}
Mosh Levy, Alon Jacoby, and Yoav Goldberg. 2024.
\newblock Same task, more tokens: the impact of input length on the reasoning performance of large language models.
\newblock In \emph{Proceedings of the 62nd Annual Meeting of the Association for Computational Linguistics (Volume 1: Long Papers)}, pages 15339--15353.

\bibitem[{Li et~al.(2024)Li, Zhang, Liu, and Chen}]{li2024needlebench}
Mo~Li, Songyang Zhang, Yunxin Liu, and Kai Chen. 2024.
\newblock Needlebench: Can llms do retrieval and reasoning in 1 million context window?
\newblock \emph{arXiv preprint arXiv:2407.11963}.

\bibitem[{Liu et~al.(2024)Liu, Wang, and Yuan}]{QuerySum}
Yushan Liu, Zili Wang, and Ruifeng Yuan. 2024.
\newblock Querysum: {A} multi-document query-focused summarization dataset augmented with similar query clusters.
\newblock In \emph{{AAAI}}, pages 18725--18732. {AAAI} Press.

\bibitem[{Ma et~al.()Ma, Zang, Chen, Chen, Jiao, Li, Lu, Liu, Ma, Dong et~al.}]{mammlongbench}
Yubo Ma, Yuhang Zang, Liangyu Chen, Meiqi Chen, Yizhu Jiao, Xinze Li, Xinyuan Lu, Ziyu Liu, Yan Ma, Xiaoyi Dong, et~al.
\newblock Mmlongbench-doc: Benchmarking long-context document understanding with visualizations.
\newblock In \emph{The Thirty-eight Conference on Neural Information Processing Systems Datasets and Benchmarks Track}.

\bibitem[{OpenAI(2024{\natexlab{a}})}]{openai2024gpt4o}
OpenAI. 2024{\natexlab{a}}.
\newblock \href {https://openai.com/index/hello-gpt-4o/} {Hello gpt-4o}.

\bibitem[{OpenAI(2024{\natexlab{b}})}]{o1card}
OpenAI. 2024{\natexlab{b}}.
\newblock \href {https://cdn.openai.com/o1-system-card-20241205.pdf} {Openai o1 system card}.

\bibitem[{Roberts et~al.()Roberts, Han, and Albanie}]{robertsneedle}
Jonathan Roberts, Kai Han, and Samuel Albanie.
\newblock Needle threading: Can llms follow threads through near-million-scale haystacks?
\newblock In \emph{The Thirteenth International Conference on Learning Representations}.

\bibitem[{Song et~al.(2024)Song, Zheng, and Luo}]{song2024counting}
Mingyang Song, Mao Zheng, and Xuan Luo. 2024.
\newblock Counting-stars: A simple, efficient, and reasonable strategy for evaluating long-context large language models.
\newblock \emph{arXiv preprint arXiv:2403.11802}.

\bibitem[{Tang et~al.(2024)Tang, Zhou, Li, Ji, Hou, and Zhang}]{tang2024citeeval}
Zecheng Tang, Keyan Zhou, Juntao Li, Baibei Ji, Jianye Hou, and Min Zhang. 2024.
\newblock L-citeeval: Do long-context models truly leverage context for responding?
\newblock \emph{arXiv preprint arXiv:2410.02115}.

\bibitem[{Team(2024)}]{qwen2.5}
Qwen Team. 2024.
\newblock \href {https://qwenlm.github.io/blog/qwen2.5/} {Qwen2.5: A party of foundation models}.

\bibitem[{Vodrahalli et~al.(2024)Vodrahalli, Ontanon, Tripuraneni, Xu, Jain, Shivanna, Hui, Dikkala, Kazemi, Fatemi et~al.}]{vodrahalli2024michelangelo}
Kiran Vodrahalli, Santiago Ontanon, Nilesh Tripuraneni, Kelvin Xu, Sanil Jain, Rakesh Shivanna, Jeffrey Hui, Nishanth Dikkala, Mehran Kazemi, Bahare Fatemi, et~al. 2024.
\newblock Michelangelo: Long context evaluations beyond haystacks via latent structure queries.
\newblock \emph{arXiv preprint arXiv:2409.12640}.

\bibitem[{Wang et~al.(2024)Wang, Chen, Cheng, Liao, Zhang, Wu, Yu, Xu, Zhang, Luo, Li, Yang, Huang, and Li}]{wang-etal-2024-leave}
Minzheng Wang, Longze Chen, Fu~Cheng, Shengyi Liao, Xinghua Zhang, Bingli Wu, Haiyang Yu, Nan Xu, Lei Zhang, Run Luo, Yunshui Li, Min Yang, Fei Huang, and Yongbin Li. 2024.
\newblock \href {https://doi.org/10.18653/v1/2024.emnlp-main.322} {Leave no document behind: Benchmarking long-context {LLM}s with extended multi-doc {QA}}.
\newblock In \emph{Proceedings of the 2024 Conference on Empirical Methods in Natural Language Processing}, pages 5627--5646, Miami, Florida, USA. Association for Computational Linguistics.

\bibitem[{Wei et~al.(2022)Wei, Wang, Schuurmans, Bosma, Xia, Chi, Le, Zhou et~al.}]{wei2022chain}
Jason Wei, Xuezhi Wang, Dale Schuurmans, Maarten Bosma, Fei Xia, Ed~Chi, Quoc~V Le, Denny Zhou, et~al. 2022.
\newblock Chain-of-thought prompting elicits reasoning in large language models.
\newblock \emph{Advances in neural information processing systems}, 35:24824--24837.

\bibitem[{Wu et~al.(2025)Wu, Yu, Liu, Yeung, and Zhou}]{wu-etal-2025-understanding}
Junjie Wu, Mo~Yu, Lemao Liu, Dit-Yan Yeung, and Jie Zhou. 2025.
\newblock \href {https://aclanthology.org/2025.naacl-long.423/} {Understanding {LLM}s' fluid intelligence deficiency: An analysis of the {ARC} task}.
\newblock In \emph{Proceedings of the 2025 Conference of the Nations of the Americas Chapter of the Association for Computational Linguistics: Human Language Technologies (Volume 1: Long Papers)}, pages 8339--8360, Albuquerque, New Mexico. Association for Computational Linguistics.

\bibitem[{Xu et~al.(2024)Xu, Li, Yu, and Zhou}]{xu2024fine}
Liyan Xu, Jiangnan Li, Mo~Yu, and Jie Zhou. 2024.
\newblock Fine-grained modeling of narrative context: A coherence perspective via retrospective questions.
\newblock In \emph{Proceedings of the 62nd Annual Meeting of the Association for Computational Linguistics (Volume 1: Long Papers)}, pages 5822--5838.

\bibitem[{Yen et~al.(2024)Yen, Gao, Hou, Ding, Fleischer, Izsak, Wasserblat, and Chen}]{yen2024helmet}
Howard Yen, Tianyu Gao, Minmin Hou, Ke~Ding, Daniel Fleischer, Peter Izsak, Moshe Wasserblat, and Danqi Chen. 2024.
\newblock Helmet: How to evaluate long-context language models effectively and thoroughly.
\newblock \emph{arXiv preprint arXiv:2410.02694}.

\bibitem[{Yu et~al.(2023)Yu, Li, Yao, Pang, Zhou, Xiao, Meng, and Zhou}]{yu2023personality}
Mo~Yu, Jiangnan Li, Shunyu Yao, Wenjie Pang, Xiaochen Zhou, Zhou Xiao, Fandong Meng, and Jie Zhou. 2023.
\newblock Personality understanding of fictional characters during book reading.
\newblock In \emph{Proceedings of the 61st Annual Meeting of the Association for Computational Linguistics (Volume 1: Long Papers)}, pages 14784--14802.

\bibitem[{Yu et~al.(2025)Yu, Liu, Wu, Chung, Zhang, Li, Yeung, and Zhou}]{yu-etal-2025-stochastic}
Mo~Yu, Lemao Liu, Junjie Wu, Tsz~Ting Chung, Shunchi Zhang, Jiangnan Li, Dit-Yan Yeung, and Jie Zhou. 2025.
\newblock \href {https://aclanthology.org/2025.naacl-long.569/} {The stochastic parrot on {LLM}`s shoulder: A summative assessment of physical concept understanding}.
\newblock In \emph{Proceedings of the 2025 Conference of the Nations of the Americas Chapter of the Association for Computational Linguistics: Human Language Technologies (Volume 1: Long Papers)}, pages 11416--11431, Albuquerque, New Mexico. Association for Computational Linguistics.

\bibitem[{Zhang et~al.(2024{\natexlab{a}})Zhang, Bai, Lv, Gu, Liu, Zou, Cao, Hou, Dong, Feng et~al.}]{zhang2024longcite}
Jiajie Zhang, Yushi Bai, Xin Lv, Wanjun Gu, Danqing Liu, Minhao Zou, Shulin Cao, Lei Hou, Yuxiao Dong, Ling Feng, et~al. 2024{\natexlab{a}}.
\newblock Longcite: Enabling llms to generate fine-grained citations in long-context qa.
\newblock \emph{arXiv preprint arXiv:2409.02897}.

\bibitem[{Zhang et~al.(2024{\natexlab{b}})Zhang, Li, Liu, Yang, Liu, Chen, Luo, and Yang}]{zhang2024marathon}
Lei Zhang, Yunshui Li, Ziqiang Liu, Jiaxi Yang, Junhao Liu, Longze Chen, Run Luo, and Min Yang. 2024{\natexlab{b}}.
\newblock Marathon: A race through the realm of long context with large language models.
\newblock In \emph{Proceedings of the 62nd Annual Meeting of the Association for Computational Linguistics (Volume 1: Long Papers)}, pages 5201--5217.

\bibitem[{Zhang et~al.(2024{\natexlab{c}})Zhang, Chen, Hu, Xu, Chen, Hao, Han, Thai, Wang, Liu et~al.}]{zhang2024bench}
Xinrong Zhang, Yingfa Chen, Shengding Hu, Zihang Xu, Junhao Chen, Moo Hao, Xu~Han, Zhen Thai, Shuo Wang, Zhiyuan Liu, et~al. 2024{\natexlab{c}}.
\newblock ∞ bench: Extending long context evaluation beyond 100k tokens.
\newblock In \emph{Proceedings of the 62nd Annual Meeting of the Association for Computational Linguistics (Volume 1: Long Papers)}, pages 15262--15277.

\bibitem[{Zheng et~al.(2024)Zheng, Zhang, Zhang, Ye, Luo, Feng, and Ma}]{zheng2024llamafactory}
Yaowei Zheng, Richong Zhang, Junhao Zhang, Yanhan Ye, Zheyan Luo, Zhangchi Feng, and Yongqiang Ma. 2024.
\newblock \href {http://arxiv.org/abs/2403.13372} {Llamafactory: Unified efficient fine-tuning of 100+ language models}.
\newblock In \emph{Proceedings of the 62nd Annual Meeting of the Association for Computational Linguistics (Volume 3: System Demonstrations)}, Bangkok, Thailand. Association for Computational Linguistics.

\end{thebibliography}

\appendix
\clearpage

\section{Complete Results}
\label{appendix:complete}

In this section, we provide the complete results for~\tref{tab:setting1} and \tref{tab:setting2} for reference.

\begin{table*}[tb]
\small
\centering
\setlength{\tabcolsep}{1.2mm}
\begin{tabular}{ll  *{6}{c} | l  *{6}{c}}
\toprule
& & \multicolumn{6}{c}{\textbf{Single}} & \multicolumn{6}{|c}{\textbf{Multi}} \\
\midrule
 & \multirow{3}{*}{\textbf{LCLM}} & \multicolumn{3}{c}{\textbf{F1}$\uparrow$} & \multicolumn{3}{c}{\textbf{Ex Acc}$\uparrow$} &
  \multicolumn{3}{c}{\textbf{F1}$\uparrow$} & \multicolumn{3}{c}{\textbf{Ex Acc}$\uparrow$} \\
\cmidrule(lr){3-5} \cmidrule(lr){6-8} \cmidrule(lr){9-11} \cmidrule(lr){12-14}
& & 8K & 16K & 24K&8K &16K&24K&8K&16K&24K &8K&16K&24K \\
\midrule
\multirow{13}{*}{Easy} 
&ProLong-8B-64K &77.50&52.96&22.02&50.00&14.00&2.00&51.62&23.39&16.44&7.00&0.00&0.00  \\
& ProLong-8B-512K &84.17&58.76&46.35&63.00&18.00&6.00&57.02&33.96&20.02&15.00&2.00&0.00  \\
& LongCite-8B &22.00&17.95&11.02&20.00&16.00&11.00&11.90&12.50&6.42&9.00&8.00&1.00  \\
& Llama-3.1-Ins-8B &83.50&70.80&65.45&59.00&20.00&13.00&64.14&44.64&30.95&17.00&5.00&2.00  \\
& Phi-3-mini &45.12&45.27&43.27&42.00&43.00&38.00&30.74&29.65&23.04&22.00&17.00&8.00 \\
& Qwen2.5-Ins-7B &59.50&62.67&58.30&59.00&61.00&54.00&42.53&34.42&30.46&33.00&18.00&13.00  \\
& Qwen2.5-Ins-72B &94.25&85.07&81.83&86.00&62.00&55.00&81.97&78.68&73.09&70.00&59.00&39.00 \\
& Llama-3.1-Ins-70B &97.67&92.67&89.63&97.00&86.00&85.00&92.80&85.17&74.47&90.00&67.00&41.00 \\
& Llama-3.3-Ins-70B &-&-&96.97&-&-&93.00&-&-&76.52&-&-&43.00 \\
& Gemini-1.5-Flash &99.67&99.67&97.77&99.00&99.00&95.00&98.34&90.83&82.16&95.00&74.00&51.00 \\
& GPT-4o mini &99.00&98.17&95.79&99.00&96.00&90.00&95.80&93.91&87.69&95.00&83.00&67.00  \\
& Gemini-1.5-Pro &-&-&96.60&-&-&93.00&97.27&95.56&89.29&95.00&86.00&67.00 \\
& GPT-4o &99.33&98.33&94.73&98.00&95.00&86.00&98.67&97.55&93.45&96.00&92.00&75.00 \\
\midrule
\multirow{13}{*}{Medium} 
& ProLong-8B-64K &70.27&36.04&15.56&37.00&6.00&0.00&59.00&25.82&19.59&15.00&0.00&0.00  \\
& ProLong-8B-512K &75.02&48.00&28.85&47.00&15.00&3.00&58.18&35.24&24.82&22.00&4.00&0.00  \\
& LongCite-8B &15.17&3.79&6.67&13.00&3.00&4.00&20.19&7.63&5.74&14.00&1.00&1.00  \\
& Llama-3.1-Ins-8B &74.37&59.32&53.99&46.00&19.00&6.00&61.13&44.42&30.00&19.00&3.00&0.00  \\
& Phi-3-mini &41.08&37.71&37.56&37.00&30.00&28.00&30.50&27.30&20.82&19.00&15.00&5.00 \\
& Qwen2.5-Ins-7B &58.50&44.67&44.29&57.00&37.00&34.00&42.34&35.29&24.63&33.00&14.00&8.00  \\
& Qwen2.5-Ins-72B &87.47&83.78&78.48&78.00&66.00&51.00&87.93&82.71&70.77&74.00&55.00&22.00 \\
& Llama-3.1-Ins-70B &90.33&87.93&79.26&89.00&78.00&61.00&93.60&78.26&64.43&90.00&46.00&19.00 \\
& Llama-3.3-Ins-70B &-&-&92.29&-&-&79.00&-&-&66.92&-&-&19.00 \\
& Gemini-1.5-Flash &99.50&98.16&95.17&99.00&94.00&86.00&95.90&88.36&74.19&88.00&64.00&29.00 \\
& GPT-4o mini &98.67&97.67&94.28&98.00&94.00&87.00&95.00&93.83&85.49&91.00&78.00&52.00  \\
& Gemini-1.5-Pro &-&-&95.88&-&-&88.00&96.50&91.40&80.20&94.00&73.00&44.00 \\
& GPT-4o &99.67&95.83&95.26&99.00&88.00&87.00&96.67&97.41&90.35&92.00&91.00&61.00 \\
\midrule
\multirow{13}{*}{Hard} 
& ProLong-8B-64K &65.72&36.75&21.71&35.00&4.00&0.00&56.70&40.55&38.11&10.00&0.00&0.00  \\
& ProLong-8B-512K &69.80&44.22&29.62&37.00&5.00&4.00&55.17&42.04&40.06&12.00&0.00&0.00  \\
& LongCite-8B &16.12&12.81&9.79&8.00&8.00&3.00&16.89&7.59&10.20&3.00&0.00&0.00  \\
& Llama-3.1-Ins-8B &73.16&57.18&50.69&33.00&11.00&5.00&60.14&45.97&38.85&12.00&2.00&0.00  \\
& Phi-3-mini &49.23&38.74&31.38&30.00&20.00&12.00&36.19&27.04&21.82&10.00&1.00&0.00 \\
& Qwen2.5-Ins-7B &61.60&47.95&36.86&49.00&26.00&15.00&43.39&31.50&20.37&14.00&1.00&0.00  \\
& Qwen2.5-Ins-72B &88.00&82.85&79.88&79.00&64.00&50.00&84.61&72.23&60.90&56.00&15.00&5.00 \\
& Llama-3.1-Ins-70B &94.20&86.33&79.12&90.00&67.00&50.00&87.38&63.87&52.21&61.00&13.00&4.00 \\
& Llama-3.3-Ins-70B &-&-&88.80&-&-&63.00&-&-&56.23&-&-&4.00 \\
& Gemini-1.5-Flash &97.80&91.86&87.60&94.00&76.00&59.00&88.46&71.22&64.65&55.00&14.00&2.00 \\
& GPT-4o mini &96.93&95.19&86.44&93.00&87.00&66.00&92.10&79.54&68.64&71.00&18.00&7.00  \\
& Gemini-1.5-Pro &-&-&93.42&-&-&77.00&93.19&74.59&65.24&74.00&17.00&9.00 \\
& GPT-4o &100.00&99.33&98.49&100.00&98.00&94.00&94.81&86.61&75.38&85.00&43.00&19.00 \\
\bottomrule
\end{tabular}
\caption{Complete results on Ref-long-A. As mentioned in~\sref{sec:setting1 results}, we do not run Llama-3.3-Ins-70B on experiments other than~\tref{tab:setting1}. Also, due to the limited budget of Gemini-1.5-Pro's API, we also not run it on experiments other than~\tref{tab:setting1} and~\fref{fig:ablation}.}
\label{tab:complete setting1}
\end{table*}

\begin{table*}[tb]
\small
\centering
\setlength{\tabcolsep}{0.4mm}
\begin{tabular}{l  l  *{7}{c} | *{7}{c}}
\toprule
\multirow{3}{*}{\textbf{Topic}} & \multirow{3}{*}{\textbf{LCLM}} &
  \multicolumn{7}{c}{\textbf{F1}$\uparrow$} & \multicolumn{7}{c}{\textbf{Ex Acc}$\uparrow$} \\
\cmidrule(lr){3-9} \cmidrule(lr){10-16} 
& & 8K & 16K & 24K & 32K & 40K & 48K & 56K & 8K & 16K & 24K & 32K & 40K & 48K & 56K \\
\midrule
\multirow{12}{*}{Foot Locker} 
& LongCite-8B &31.86&15.71&10.26&12.27&8.61&9.44&8.01&6.00&0.00&0.00&0.00&0.00&0.00&0.00  \\
& Qwen2.5-Ins-7B &61.51&48.46&41.36&35.60&34.22&29.27&25.85&26.00&7.00&3.00&2.00&1.00&0.00&0.00  \\
& Phi-3-mini &59.12&46.66&45.16&32.11&27.52&24.77&20.98&36.00&12.00&9.00&4.00&2.00&2.00&1.00 \\
& ProLong-8B-64K &77.68&55.49&42.90&38.59&27.25&23.95&18.00&49.00&7.00&3.00&2.00&0.00&0.00&0.00  \\
& ProLong-8B-512K &78.25&60.30&42.96&40.68&35.67&30.04&27.02&52.00&13.00&3.00&0.00&0.00&0.00&0.00  \\
& Llama-3.1-Ins-8B &86.81&67.39&55.76&49.90&41.28&31.63&25.99&65.00&25.00&9.00&6.00&4.00&2.00&0.00  \\
& Qwen2.5-Ins-72B &88.84&82.91&77.82&73.24&73.75&70.35&66.81&66.00&52.00&32.00&20.00&19.00&16.00&13.00 \\
& GPT-4o mini &88.51&79.54&77.49&73.09&69.82&66.90&62.33&69.00&45.00&29.00&19.00&17.00&14.00&11.00  \\
& Llama-3.1-Ins-70B &91.22&85.35&76.75&69.56&60.31&55.10&52.02&72.00&52.00&35.00&17.00&10.00&5.00&3.00 \\
& Gemini-1.5-Flash &93.29&86.73&80.55&76.21&75.89&73.98&72.01&82.00&61.00&44.00&27.00&19.00&16.00&9.00 \\
& Gemini-1.5-Pro &-&-&-&-&-&-&-&-&-&-&-&-&-&- \\
& GPT-4o &91.10&87.39&84.84&84.64&81.68&80.71&79.86&74.00&62.00&49.00&40.00&30.00&28.00&26.00 \\
\midrule
\multirow{12}{*}{Twitter} 
& LongCite-8B &20.90&10.47&12.45&7.92&9.99&7.26&7.80&4.00&0.00&0.00&0.00&0.00&0.00&0.00  \\
& Qwen2.5-Ins-7B &49.98&38.81&39.21&33.59&29.77&27.04&24.87&16.00&1.00&0.00&0.00&0.00&0.00&0.00  \\
& Phi-3-mini &53.72&37.84&33.54&28.42&26.11&23.92&22.92&26.00&4.00&0.00&0.00&0.00&0.00&0.00 \\
& ProLong-8B-64K &70.02&52.36&45.66&36.83&30.11&26.80&24.06&33.00&5.00&3.00&0.00&0.00&0.00&0.00  \\
& ProLong-8B-512K &73.84&52.39&42.75&37.40&31.10&29.42&25.95&39.00&6.00&1.00&0.00&0.00&0.00&0.00  \\
& Llama-3.1-Ins-8B &81.86&63.05&52.37&41.70&37.19&35.21&30.53&52.00&14.00&7.00&2.00&2.00&0.00&0.00  \\
& Qwen2.5-Ins-72B &83.43&76.35&74.92&71.30&68.00&64.87&62.00&55.00&34.00&19.00&10.00&6.00&0.00&1.00 \\
& GPT-4o mini &89.85&80.53&76.67&71.30&63.48&60.73&55.71&70.00&39.00&22.00&15.00&4.00&3.00&1.00  \\
& Llama-3.1-Ins-70B &91.87&86.21&80.08&72.01&61.43&54.35&48.25&77.00&54.00&34.00&18.00&8.00&4.00&4.00 \\
& Gemini-1.5-Flash &89.97&86.05&79.45&76.47&74.27&72.74&70.85&77.00&52.00&27.00&15.00&11.00&10.00&6.00 \\
& Gemini-1.5-Pro &90.10&84.50&80.47&76.74&68.33&62.80&61.20&78.00&56.00&39.00&28.00&15.00&12.00&10.00 \\
& GPT-4o &90.00&84.93&83.50&80.85&80.43&75.96&73.81&71.00&56.00&41.00&32.00&20.00&11.00&8.00 \\
\midrule
\multirow{12}{*}{Financial Market} 
& LongCite-8B &31.45&19.90&13.38&11.51&8.65&11.58&13.91&6.00&0.00&0.00&0.00&0.00&0.00&0.00  \\
& Phi-3-mini &49.41&37.64&33.95&29.97&25.40&23.24&23.11&16.00&2.00&1.00&0.00&0.00&0.00&0.00 \\
& Qwen2.5-Ins-7B &56.18&39.44&34.25&31.67&28.24&24.38&25.00&25.00&6.00&3.00&0.00&1.00&0.00&0.00  \\
& ProLong-8B-64K &71.68&50.08&41.84&33.58&27.04&24.88&20.34&29.00&4.00&2.00&0.00&0.00&0.00&0.00  \\
& ProLong-8B-512K &72.01&52.59&42.20&36.98&30.44&29.61&27.26&31.00&7.00&1.00&1.00&0.00&0.00&0.00  \\
& Llama-3.1-Ins-8B &77.90&64.77&47.44&40.27&34.88&30.69&28.65&43.00&16.00&2.00&1.00&1.00&0.00&0.00  \\
& Qwen2.5-Ins-72B &90.06&83.62&75.52&75.09&73.09&67.22&66.64&69.00&43.00&18.00&19.00&12.00&7.00&5.00 \\
& GPT-4o mini &91.05&84.41&76.37&69.93&64.63&57.49&53.27&74.00&41.00&22.00&18.00&8.00&3.00&1.00  \\
& Llama-3.1-Ins-70B &93.47&82.36&77.84&66.04&54.48&48.90&38.96&79.00&39.00&27.00&12.00&3.00&2.00&0.00 \\
& Gemini-1.5-Flash &94.20&88.46&83.73&80.59&77.06&72.50&71.69&80.00&54.00&37.00&21.00&11.00&7.00&6.00 \\
& Gemini-1.5-Pro &-&-&-&-&-&-&-&-&-&-&-&-&-&- \\
& GPT-4o &94.41&90.78&87.04&83.75&81.80&78.83&73.24&81.00&65.00&46.00&36.00&30.00&18.00&13.00 \\
\bottomrule
\end{tabular}
\caption{Complete results on Ref-long-F. Due to the limited budget of Gemini-1.5-Pro's API, we also not run it on experiments other than~\tref{tab:setting2}.}
\label{tab:complete setting2}
\end{table*}

\section{Prompts We Used in This Work}
\label{appendix:prompts}
For each experimental setting, we evaluated two prompt design methods: (A) providing questions after the documents and (B) providing questions before the documents. We observed that different models demonstrated preferences for distinct prompt designs. For instance, LLama-3.1-Ins-8B performed better with prompt design (A), whereas LLama-3.1-Ins-70B favored design (B). To ensure each model operated at its full potential, we evaluated both prompt design methods for all the LCLMs on Ref-Long-A to identify the optimal design for each LCLM. In the rest two subsets, we applied the best-performing prompt design for each model. Additionally, for LongCite-8B, which follows a specific prompt template, we placed the documents in its context variable and put the description and instructions in the query variable to align with its template requirements. 

\fref{fig:prompt1_ori}, \fref{fig:prompt1_before}, \fref{fig:prompt2_ori}, \fref{fig:prompt2_before}, \fref{fig:prompt3_ori}, and \fref{fig:prompt3_before} show the two prompt design on three subsets of Ref-Long.

\paragraph{Prompts with human strategy.} \fref{fig:prompt1_ori_strategy} and \fref{fig:prompt1_before_strategy} list the prompts with human strategy used in~\sref{sec:factors}.

\paragraph{Prompts with natural language.} \fref{fig:prompt1_ori_fruit} and \fref{fig:prompt1_before_fruit} list the prompts with human strategy used in~\sref{sec:factors}. We also list the mapping between number$\in [0, 20)$ and 20 fruit names in~\tref{tab:mapping}. 

\begin{figure}[h]
\begin{tcolorbox}[colback=black!7.5!white, colframe=black!80!white, title=Example Prompt (A), fontupper=\footnotesize, fonttitle=\footnotesize]

\texttt{[User Input]}: \vspace{2pt}\\
You will find several documents indexed with numbers below. Each document contains one or more sentences describing a little penguin collecting specific numbers of stars in the format of: The little penguin counted \{num\} \ding{72} ". Please read through these documents carefully and answer some questions. \\ \newline
0: \{Document 0\} \\ 
1: \{Document 1\} \\ 
2: \{Document 2\} \\ 

Could you tell me the indexes of all documents where the little penguin counts \texttt{88} stars? Please provide your answer in the following format without explanations: "Documents: \{\}". \\

\end{tcolorbox}

\caption{Example Prompt for Ref-Long-A: Instructions Followed by Documents, where ``\texttt{88}'' is the interested key.}
\label{fig:prompt1_ori}
\end{figure}

\begin{figure}[h]
\begin{tcolorbox}[colback=black!7.5!white, colframe=black!80!white, title=Example Prompt (B) , fontupper=\footnotesize, fonttitle=\footnotesize]

\texttt{[User Input]}: \vspace{2pt}\\
You will find several documents indexed with numbers below. Each document contains one or more sentences describing a little penguin collecting specific numbers of stars in the format of: "The little penguin counted \{num\} \ding{72}". Please read through these documents carefully and answer my question: could you tell me the indexes of all documents where the little penguin counts \texttt{88} stars? Please provide your answer in the following format without explanations: "Documents: \{\} ". \\
\newline
0: \{Document 0\} \\ 
1: \{Document 1\} \\ 
2: \{Document 2\} \\ 
\end{tcolorbox}

\caption{Example Prompt for Ref-Long-A: Instructions Preceding Documents, where ``\texttt{88}'' is the interested key.}
\label{fig:prompt1_before}
\end{figure}

\begin{figure}[h]
\begin{tcolorbox}[colback=black!7.5!white, colframe=black!80!white, title=Example Prompt (A), fontupper=\footnotesize, fonttitle=\footnotesize]

\texttt{[User Input]}: \vspace{2pt}\\
You will find several documents indexed with numbers below. Each document contains one or more insights (statements that contain specific information about the given topic 'Strategic Growth and Operational Changes at Foot Locker') in the format of sentences. Please read through these documents carefully and answer some questions. \\
\newline
0: \{Document 0\} \\ 
1: \{Document 1\} \\ 
2: \{Document 2\} \\ 

Could you tell me the indexes of all documents talking about the insight "\texttt{Foot Locker enhances its loyalty program to include invitations to special events at flagship stores, such as guest appearances by NBA players, starting in the 2023 holiday season.}"? Please provide your answer in the following format without explanations: "Documents: \{\}".
\end{tcolorbox}

\caption{Example Prompt for Ref-Long-F: Instructions Followed by Documents, where ``\texttt{Foot Locker enhances its loyalty program to include invitations to special events at flagship stores, such as guest appearances by NBA players, starting in the 2023 holiday season.}'' is the interested key.}
\label{fig:prompt2_ori}
\end{figure}

\begin{figure}[h]
\begin{tcolorbox}[colback=black!7.5!white, colframe=black!80!white, title=Example Prompt (B), fontupper=\footnotesize, fonttitle=\footnotesize]

\texttt{[User Input]}: \vspace{2pt}\\
You will find several documents indexed with numbers below. Each document contains one or more insights (statements that contain specific information about the given topic 'Strategic Growth and Operational Changes at Foot Locker') in the format of sentences. Please read through these documents carefully and answer my question: could you tell me the indexes of all documents talking about the insight "\texttt{Foot Locker enhances its loyalty program to include invitations to special events at flagship stores, such as guest appearances by NBA players, starting in the 2023 holiday season.}"? Please provide your answer in the following format without explanations: "Documents: \{\}".\\
\newline
0: \{Document 0\} \\ 
1: \{Document 1\} \\ 
2: \{Document 2\} \\ 
\end{tcolorbox}

\caption{Example Prompt for Ref-Long-F: Instructions Preceding Documents, where ``\texttt{Foot Locker enhances its loyalty program to include invitations to special events at flagship stores, such as guest appearances by NBA players, starting in the 2023 holiday season.}'' is the interested key.}
\label{fig:prompt2_before}
\end{figure}

\begin{figure}[h]
\begin{tcolorbox}[colback=black!7.5!white, colframe=black!80!white, title=Example Prompt (A), fontupper=\footnotesize, fonttitle=\footnotesize]

\texttt{[User Input]}: \vspace{2pt}\\
You will find several papers indexed with numbers below. Each paper is divided by  ``\verb|\n\n|''. Please read through these papers carefully and answer some questions. \\
\newline
0: \{Paper 0\} \\ 
1: \{Paper 1\} \\ 
2: \{Paper 2\} \\ 

Could you tell me the indexes of all papers citing "\texttt{ClimODE: Climate and Weather Forecasting with Physics-informed Neural ODEs}"? Please provide your answer in the following format without explanations: "Papers: \{\}".
\end{tcolorbox}

\caption{Example Prompt for Ref-Long-Paper: Instructions Followed by Documents, where ``\texttt{ClimODE: Climate and Weather Forecasting with Physics-informed Neural ODEs}'' is the specific key.}
\label{fig:prompt3_ori}
\end{figure}

\begin{figure}[h]
\begin{tcolorbox}[colback=black!7.5!white, colframe=black!80!white, title=Example Prompt (B), fontupper=\footnotesize, fonttitle=\footnotesize]

\texttt{[User Input]}: \vspace{2pt}\\
You will find several papers indexed with numbers below. Each paper is divided by  ``\verb|\n\n|''. Please read through these papers carefully and answer my question: could you tell me the indexes of all papers citing ``\texttt{ClimODE: Climate and Weather Forecasting with Physics-informed Neural ODEs}''? Please provide your answer in the following format without explanations: Papers: \{\}.\\
\newline
0: \{Paper 0\} \\ 
1: \{Paper 1\} \\ 
2: \{Paper 2\} \\ 
\end{tcolorbox}

\caption{Example Prompt for Ref-Long-Paper: Instructions Preceding Documents,  where "\texttt{ClimODE: Climate and Weather Forecasting with Physics-informed Neural ODEs}" is the specific key.}
\label{fig:prompt3_before}
\end{figure}

\begin{figure}[h]
\begin{tcolorbox}[colback=black!7.5!white, colframe=black!80!white, title=Example Prompt (A), fontupper=\footnotesize, fonttitle=\footnotesize]

\texttt{[User Input]}: \vspace{2pt}\\
You will find several documents indexed with numbers below. Each document contains one or more sentences describing a little penguin collecting specific numbers of stars in the format of: "The little penguin counted \{num\} \ding{72} ". Please read through these documents carefully while do the following:\\
1. Create an empty Python dictionary called star\_dict.\\
2. When you encounter a sentence in document K of the form "The little penguin counted \{num\} \ding{72}":\\
- If num is not already a key in star\_dict, add it with star\_dict[num] = [K].\\
- If num is already a key, append the document identifier K to the list with star\_dict[num].append(K).\\
After processing the documents, use the populated star\_dict to answer some questions. \\ \newline
0: \{Document 0\} \\ 
1: \{Document 1\} \\ 
2: \{Document 2\} \\ 

Could you tell me the indexes of all documents where the little penguin counts \texttt{88} stars? Please provide your answer in the following format without explanations: "Documents: \{\}". \\

\end{tcolorbox}

\caption{Example Prompt for Ref-Long-A: Instructions Followed by Documents, with human strategy, where ``\texttt{88}'' is the interested key.}
\label{fig:prompt1_ori_strategy}
\end{figure}

\begin{figure}[h]
\begin{tcolorbox}[colback=black!7.5!white, colframe=black!80!white, title=Example Prompt (B) , fontupper=\footnotesize, fonttitle=\footnotesize]

\texttt{[User Input]}: \vspace{2pt}\\
You will find several documents indexed with numbers below. Each document contains one or more sentences describing a little penguin collecting specific numbers of stars in the format of: "The little penguin counted \{num\} \ding{72}". Please read through these documents carefully while do the following:\\
1. Create an empty Python dictionary called star\_dict.\\
2. When you encounter a sentence in document K of the form "The little penguin counted \{num\} \ding{72}":\\
- If num is not already a key in star\_dict, add it with star\_dict[num] = [K].\\
- If num is already a key, append the document identifier K to the list with star\_dict[num].append(K).\\
After processing the documents, use the populated star\_dict to answer my question: could you tell me the indexes of all documents where the little penguin counts \texttt{88} stars? Please provide your answer in the following format without explanations: "Documents: \{\} ". \\
\newline
0: \{Document 0\} \\ 
1: \{Document 1\} \\ 
2: \{Document 2\} \\ 
\end{tcolorbox}

\caption{Example Prompt for Ref-Long-A: Instructions Preceding Documents, with human strategy, where ``\texttt{88}'' is the interested key.}
\label{fig:prompt1_before_strategy}
\end{figure}

\begin{figure}[h]
\begin{tcolorbox}[colback=black!7.5!white, colframe=black!80!white, title=Example Prompt (A), fontupper=\footnotesize, fonttitle=\footnotesize]

\texttt{[User Input]}: \vspace{2pt}\\
You will find several documents indexed with numbers below. Each document contains one or more sentences describing a little penguin eating a specific fruit in the format of: The little penguin eated \{fruit\}". Please read through these documents carefully and answer some questions. \\ \newline
0: \{Document 0\} \\ 
1: \{Document 1\} \\ 
2: \{Document 2\} \\ 

Could you tell me the indexes of all documents where the little penguin eats \texttt{apple}? Please provide your answer in the following format without explanations: "Documents: \{\}". \\

\end{tcolorbox}

\caption{Example Prompt for Ref-Long-A: Instructions Followed by Documents, with fruit name instead, where ``\texttt{apple}'' is the interested key.}
\label{fig:prompt1_ori_fruit}
\end{figure}

\begin{figure}[h]
\begin{tcolorbox}[colback=black!7.5!white, colframe=black!80!white, title=Example Prompt (B) , fontupper=\footnotesize, fonttitle=\footnotesize]

\texttt{[User Input]}: \vspace{2pt}\\
You will find several documents indexed with numbers below. Each document contains one or more sentences describing a little penguin eating a specific fruit in the format of: The little penguin eated \{fruit\}". Please read through these documents carefully and answer my question: could you tell me the indexes of all documents where the little penguin eats \texttt{apple}? Please provide your answer in the following format without explanations: "Documents: \{\} ". \\
\newline
0: \{Document 0\} \\ 
1: \{Document 1\} \\ 
2: \{Document 2\} \\ 
\end{tcolorbox}

\caption{Example Prompt for Ref-Long-A: Instructions Followed by Documents, with fruit name instead, where ``\texttt{apple}'' is the interested key.}
\label{fig:prompt1_before_fruit}
\end{figure}

\begin{figure}[h]
\begin{tcolorbox}[colback=black!7.5!white, colframe=black!80!white, title=Example Prompt (A), fontupper=\footnotesize, fonttitle=\footnotesize]

\texttt{[User Input]}: \vspace{2pt}\\
You will find several documents indexed with numbers below. Each document is divided by '\textbackslash n\textbackslash n'. Please read through these documents carefully and answer some questions. \\
\newline
0: \{Document 0\} \\ 
1: \{Document 1\} \\ 
2: \{Document 2\} \\ 

Could you tell me the indexes of all documents discussing "\texttt{Paul George}"? Please provide your answer in the following format without explanations: "Documents: \{\}".
\end{tcolorbox}

\caption{Example Prompt for Ref-Long-NBA: Instructions Followed by Documents, where ``\texttt{Paul George}'' is the interested key.}
\label{fig:prompt4_ori}
\end{figure}

\begin{figure}[h]
\begin{tcolorbox}[colback=black!7.5!white, colframe=black!80!white, title=Example Prompt (B), fontupper=\footnotesize, fonttitle=\footnotesize]

\texttt{[User Input]}: \vspace{2pt}\\
You will find several documents indexed with numbers below. Each document is divided by '\textbackslash n\textbackslash n'. Please read through these documents carefully and answer my question: could you tell me the indexes of all documents discussing "\texttt{Paul George}"? Please provide your answer in the following format without explanations: "Documents: \{\}".\\
\newline
0: \{Document 0\} \\ 
1: \{Document 1\} \\ 
2: \{Document 2\} \\ 
\end{tcolorbox}

\caption{Example Prompt for Ref-Long-NBA: Instructions Preceding Documents, where ``\texttt{Paul George}'' is the interested key.}
\label{fig:prompt4_before}
\end{figure}

\begin{table}[h!]
\centering
\begin{tabular}{|c|l|}
\hline
%\textbf{Key} & \textbf{Value} \\ \hline
0  & apple        \\ \hline
1  & banana       \\ \hline
2  & orange       \\ \hline
3  & mango        \\ \hline
4  & grapes       \\ \hline
5  & pineapple    \\ \hline
6  & strawberry   \\ \hline
7  & blueberry    \\ \hline
8  & raspberry    \\ \hline
9  & watermelon   \\ \hline
10 & papaya       \\ \hline
11 & kiwi         \\ \hline
12 & peach        \\ \hline
13 & pear         \\ \hline
14 & cherry       \\ \hline
15 & plum         \\ \hline
16 & dragonfruit  \\ \hline
17 & pomegranate  \\ \hline
18 & lychee       \\ \hline
19 & fig          \\ \hline
\end{tabular}
\caption{Mapping between numbers and fruit names.}
\label{tab:mapping}
\end{table}

\section{Fluent Key in Fluent Context Setting on Real-world Data}
\label{appendix:nba data}
Since the Ref-Long-F dataset in~\sref{sec:fluent with fluent} is constructed from synthetic documents, we further develop a real-world counterpart named \textbf{Ref-Long-F-NBA} under the same setting to validate our findings. Specifically, we collect 47 documents from the internet, each containing approximately 1,200 tokens and focusing exclusively on a single NBA player. Note that multiple documents may exist for the same player. To construct a Ref-Long task with $\boldsymbol{M}$ documents, we first randomly select an NBA player, then use the $\boldsymbol{m}$ documents that discuss this player along with $\boldsymbol{M} - \boldsymbol{m}$ randomly sampled distractor documents to form the input. Following this procedure and due to the limited API credits, we only set $\boldsymbol{M} = 20$ to create 100 tasks, each containing around 24K input tokens.
During evaluation, LCLMs are asked to identify the indexes of documents that reference the target player, using the player's name as the query key (see prompts in Appendix~\sref{appendix:prompts}). This setup ensures that the key is fluently embedded within the original documents, aligning with our intended task setting.

We evaluate LCLMs on this dataset and report the results in~\tref{tab:nba_results} (check the prompt we use in Appendix~\sref{appendix:prompts}. Consistent with~\tref{tab:setting2}, all models exhibit significant performance gaps even with a relatively moderate input length of 24K tokens. The performance trends closely mirror those observed in~\tref{tab:setting2}, further reinforcing that the conclusions from~\sref{sec:fluent with fluent} generalize to real-world data.

\begin{table}[tb]
\renewcommand\arraystretch{1.1}
\centering
\setlength{\tabcolsep}{3mm}
\small
\begin{tabular}{lcc}
\toprule[1pt]
\textbf{LCLM} & \textbf{F1}$\uparrow$ & \textbf{Ex Acc}$\uparrow$ \\
\midrule[0.5pt]
%LongCite-8B &4.70 &0.00 \\
Qwen2.5-Ins-7B &32.85 &2.00 \\
%Phi-3-mini & & \\
%ProLong-8B-64K &32.52 &0.00 \\
ProLong-8B-512K & 39.49&2.00 \\
Llama-3.1-Ins-8B &53.58 &4.00 \\
Gemini-1.5-Flash &77.65 &16.00 \\
Qwen2.5-Ins-72B &80.60 &26.00 \\
GPT-4o mini & 81.43&28.00 \\
Llama-3.1-Ins-70B &82.40 &\textbf{32.00} \\
GPT-4o &\textbf{83.18} & 31.00\\
%Gemini-1.5-Pro & & \\
\bottomrule[1pt]
\end{tabular}
\caption{Results on the Ref-Long-F-NBA dataset with 24K input tokens. We report results only for models with an Ex Acc score greater than 0. The best scores in each column are boldfaced. Due to limited API budget, Gemini-1.5-Pro was not evaluated on this dataset.
}
%\vspace{-0.1in}
\label{tab:nba_results}
\end{table}

\section{Details of LCLM Inference}
\label{appendix:inference details}
We call APIs for commercial LLMs and perform inference for open-source LLMs on two NVIDIA A800 GPUs. Inference is conducted using vLLM~\cite{kwon2023efficient} with greedy decoding. For LLaMA-3.1-70B-Instruct and Qwen-2.5-72B-Instruct, we employ the INT4 quantized version using AWQ for inference. During inference, we set temperature to 0 and top P value to 1 to eliminate randomness and keep other hyperparameters default.

\section{Details of Fine-Tuning}
We fine-tune LLaMA-3.1-8B-Instruct using LLaMA-Factory~\cite{zheng2024llamafactory}, setting the number of epochs to 1 and the learning rate to 1e-4. We adopt a LoRA strategy~\cite{hulora} instead of training with full parameters, as we observe the latter method often leads to the model outputting the same answer for all test samples. Fine-tuning is performed on two NVIDIA A800 GPUs, taking roughly 15 minutes.
\label{appendix:fine-tune details}

\end{document}